\pdfoutput=1

\documentclass[11pt]{article}

\usepackage[final]{acl}

\usepackage{times}
\usepackage{latexsym}
\usepackage{hyperref}

\usepackage[T1]{fontenc}

\usepackage[utf8]{inputenc}

\usepackage{microtype}

\usepackage{inconsolata}

\usepackage{graphicx}
\usepackage{booktabs}
\usepackage{multirow}
\usepackage{subcaption}
\usepackage{makecell}
\usepackage{caption}
\usepackage{float}
\usepackage{amsmath}
\usepackage{xcolor} 
\usepackage{tabularx}
\usepackage{tcolorbox}
\usepackage{minted}
\usepackage{algorithm}
\usepackage{algpseudocode}
\usepackage{xspace}
\usepackage{stfloats}
\usepackage{longtable}
\usepackage[nameinlink]{cleveref}
\usepackage[normalem]{ulem}
\useunder{\uline}{\ul}{}

\title{From Chat Logs to Collective Insights:\\ Aggregative Question Answering}

\author{Wentao Zhang \\
  University of Waterloo \\
  \texttt{w564zhan@uwaterloo.ca} \\\And
  Woojeong Kim \\
  Cornell University \\
   \texttt{wk247@cornell.edu} \\\And
  Yuntian Deng \\
   University of Waterloo \\
  \texttt{yuntian@uwaterloo.ca} \\}

\begin{document}
\maketitle
\begin{abstract}
Conversational agents powered by large language models (LLMs) are rapidly becoming integral to our daily interactions, generating unprecedented amounts of conversational data. Such datasets offer a powerful lens into societal interests, trending topics, and collective concerns. Yet, existing approaches typically treat these interactions as independent and miss critical insights that could emerge from aggregating and reasoning across large-scale conversation logs. In this paper, we introduce Aggregative Question Answering, a novel task requiring models to reason over thousands of user-chatbot interactions to answer aggregative queries, such as identifying emerging concerns among specific demographics. To enable research in this direction, we constructed WildChat-AQA, a benchmark comprising 6,027 aggregative questions derived from 182,330 real-world chatbot conversations. Experiments show that existing methods either struggle to reason effectively or incur prohibitive computational costs, underscoring the need for new approaches capable of extracting collective insights from large-scale conversational data.

\end{abstract}

\section{Introduction}

Rapid adoption of conversation agents powered by large language models (LLMs) is transforming human-computer interactions, integrating deeply into society, and generating unprecedented volumes of conversational data~\citep{backlinko_chatgpt_stats_2025, DeVynck_2023}. Platforms using LLM-based chatbots now routinely handle millions of interactions every day, producing rich datasets that capture real-time dialogues reflecting user interests, emerging societal trends, and collective concerns~\citep{zhao2024wildchat, zheng2024lmsyschatm}. Such conversational data offer immense potential for deriving insights at scale, revealing patterns in societal dynamics, shifts in public sentiment, and demographic-specific concerns~\citep{valdez2020social}.

  
\begin{figure}[!t]
    \centering
    \includegraphics[trim={20px 20px 20px 20px},clip, width=0.99\linewidth]{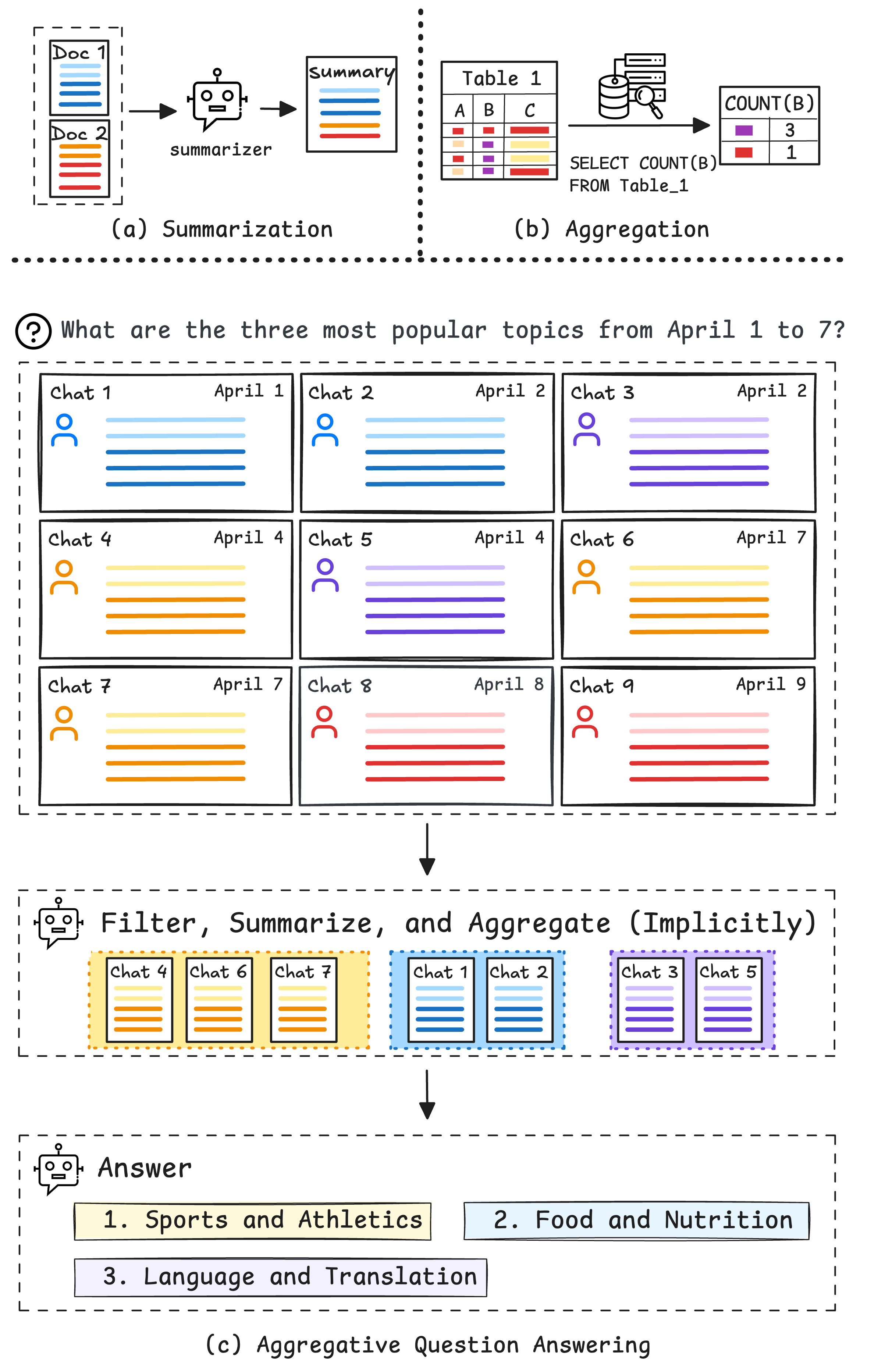}
    \caption{Comparison of different aggregation paradigms: (a) summarization, (b) aggregation over structured databases, and (c) aggregation over large sets of conversations (our focus).}    \label{fig:intro}
\end{figure}


Despite the inherent richness of these conversational datasets, current research typically treats interactions as isolated, independent data points, primarily using them to finetune LLMs for generating improved individual responses~\citep{vicuna2023, lambert2025tulu3pushingfrontiers, zhang2025bestinstructiontuningdatafit}. This independent and identically distributed (i.i.d.) assumption overlooks important temporal patterns and thematic connections that naturally arise from large-scale, real-world user-chatbot conversations. Conversations do not occur in isolation, but rather within specific temporal, geographical, and device-related contexts~\citep{tamkin2024clioprivacypreservinginsightsrealworld}. These contextual features carry significant potential for deriving collective insights, such as understanding regional differences in user concerns or identifying temporal shifts in societal attitudes---insights which are lost under the simplifying i.i.d. assumption.


To address this gap, we introduce a new task, Aggregative Question Answering, which requires reasoning across large-scale collections of user-chatbot interactions to extract aggregative insights. Unlike traditional summarization, which condenses information from one or a few documents into static summaries, Aggregative Question Answering generates dynamic answers that depend on the specific aggregative query posed. The task requires reasoning over thousands of conversations to answer questions such as identifying trending topics within specific timeframes (\textit{``What topics trended last week?''}), emerging concerns among particular demographics (\textit{``What topics are Californians concerned about before an election?''}), or tracking changes in societal sentiment (\textit{``How have users' attitudes toward AI evolved this month?''}). The core challenge thus lies not in summarizing individual conversations, but rather in global-scale reasoning conditioned on the query. \Cref{fig:intro} highlights the high-level distinctions between traditional summarization, querying structured databases, and aggregative question answering.

\begin{figure*}[h]
    \centering
    \includegraphics[width=0.94\linewidth]{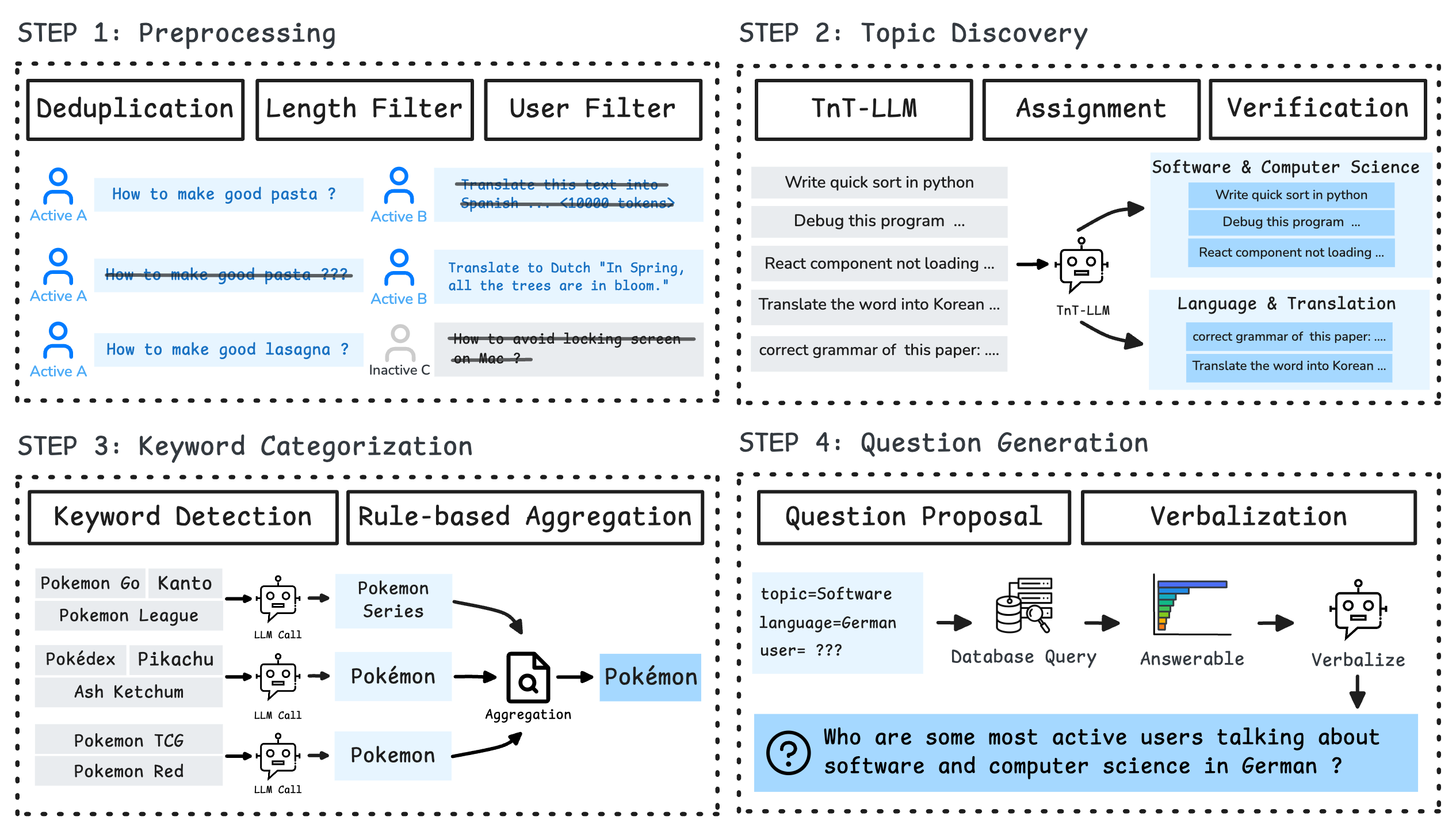}
    \caption{Overview of the WildChat-AQA dataset creation process.}
    \label{fig:data_creation}
\end{figure*}

To facilitate research into Aggregative Question Answering, we introduce WildChat-AQA, a benchmark constructed from the WildChat dataset~\citep{zhao2024wildchat,deng-etal-2024-wildvis}. WildChat captures not only conversation transcripts but also metadata such as temporal, geographical, and user-specific information. WildChat-AQA formulates aggregational queries about both explicit and implicit attributes of conversations, including topics, keywords, geographical locations, and temporal information, in a multi-choice format. A concrete example of the data creation process is shown in \Cref{fig:data_creation}. The benchmark includes 6,027 aggregative questions derived from 182,330 real-world user-chatbot conversations, reflecting genuine user interests and societal trends, thus providing a resource for evaluating models' ability to perform aggregative reasoning at scale.

We evaluated current methods, including both non-reasoning and reasoning models, adapted to this task via fine-tuning, retrieval-augmented generation (RAG), and a customized retrieval approach developed specifically for aggregative reasoning: {PROBE} ({P}robing {R}etrieval {O}f {B}road {E}vidence). Experimental results show substantial limitations in existing methods: current systems either struggle to reason effectively at scale or incur prohibitive computational costs. Even when whole oracle contexts relevant to a query are provided, there remains significant room for improvement. In more realistic settings with no access to oracle contexts, the performance drops further.

Our findings show that we need more scalable and effective methods capable of extracting collective insights from large-scale conversational datasets. While Aggregative Question Answering opens promising avenues for real-world analytics, we acknowledge potential societal impacts, particularly when insights relate to sensitive topics such as elections, public opinion, or public health. However, we believe that transparent, open academic research fosters responsible development and deployment of such powerful technologies. By introducing Aggregative Question Answering as a new task, we aim to spur future methods that fully harness the potential of large-scale conversational data, ultimately enabling deeper societal understanding and more impactful applications of LLMs.

Our benchmark, code, and dataset are publicly available at \url{https://github.com/yuntian-group/wildchat_aggregative_question_answering}, and we also provide a user-friendly benchmark visualization tool at \url{https://aggregativeqa.com/dataview}.

\section{Aggregative Question Answering}
To support research on Aggregative Question Answering, we construct the WildChat-AQA benchmark based on the WildChat dataset~\citep{zhao2024wildchat,deng-etal-2024-wildvis}. WildChat provides real-world conversations between users and chatbots, along with basic metadata such as timestamps and user locations. In this work, we extend these attributes by introducing additional attributes, such as topics and keywords inferred from the conversation text using LLMs. These inferred attributes serve as the ground truth annotations for building our benchmark. At evaluation time, models must infer them from conversations to answer aggregative questions. \Cref{tab:att_intro} summarizes the attributes, indicating which require inference and which can be obtained directly.

\begin{table}[!t]
    \centering
    \small
    \begin{tabular}{@{}l@{ }cc@{ }c@{}}
        \toprule
        \textbf{Name} & \textbf{Multi-Val} & \textbf{Inferred} & \textbf{Examples} \\
        \midrule
         Location & No &   No & United States, Canada \\
         \midrule
         User Name & No &  No & lostclasp37, toughcue8\\
         \midrule
         Time & No  & No & 4/26/2023, 1:47:24 PM \\
         \midrule
         Language & No  & No & English, Russian \\
         \midrule
         Topic & Yes & Yes & \makecell{Software, Programming \\ and Computer Science} \\
         \midrule
         Subtopic & Yes & Yes& \makecell{Mobile Development, \\AI and ML} \\ 
         \midrule
         Keywords & Yes & Yes & C++, 
Pokémon \\
         \bottomrule
    \end{tabular}
    \caption{Attributes used in WildChat-AQA. \textbf{Multi-Val} indicates whether an attribute can have multiple values per conversation. \textbf{Inferred} indicates whether the attribute must be inferred from conversation content (as opposed to being directly available from metadata). \textbf{Examples} show representative attribute values.}
    \label{tab:att_intro}
\end{table}
\subsection{Dataset Construction}

The construction of WildChat-AQA involve four main steps, as illustrated in \Cref{fig:data_creation}:

\paragraph{Step 1: Preprocessing} We begin by performing minHash-based deduplication~\citep{huggingface_dedup_2023} to remove highly similar conversations to ensure diversity. We also filter conversations that exceed 4,096 tokens to maintain manageable context lengths. Additionally, we retain only active users with at least 10 interactions to ensure sufficient user-specific data. We also generate user IDs from IP addresses and headers.


\paragraph{Step 2: Topic Discovery}  To support meaningful aggregative queries, we prompt GPT-4o to summarize each conversation and extract relevant keywords. Using these summaries, we recursively apply TnT-LLM~\citep{wan2024tntllmtextminingscale} to infer hierarchical topics at two levels: coarse-grained topics and fine-grained subtopics. Detailed prompts and examples can be found in \Cref{appendix:data_creation}.

\paragraph{Step 3: Keyword Categorization}  Certain subtopics, such as ``Programming'' and ``Fan-fiction and Crossover,'' contain many conversations. To support finer-grained aggregative queries, we further categorize keywords inferred from conversations into higher-level categories using LLMs so that we can derive aggregative information. For example, different Pokémon-related keywords (versions, characters, trademarks) are grouped into a single category ``Pokémon''. Full details of this procedure are also available in \Cref{appendix:data_creation}.


\paragraph{Step 4: Question Generation} Finally, we generate aggregative questions using combinations of attributes stored in our constructed database. This database is built by compiling all conversations along with their inferred attributes (such as topics and keywords extracted by GPT-4o) and provided metadata attributes (such as timestamps and locations). We then sample attribute combinations from zero to three attributes as conditions and a target attribute to query our database. These structured queries explicitly specify the conditions (attribute-value constraints, e.g., \textit{user=abcd}, \textit{keyword=efgh}) and the target attributes to query. The exact combinations are detailed in \Cref{tab:data_question_stat} and \Cref{sec:appendix_data_profile}. They serve two purposes: (1) retrieving ground truth answers by converting them directly into database queries executed against our database to retrieve and rank candidate answers, and (2) generating corresponding natural language questions using GPT-4.1. The prompts used are provided in \Cref{fig:generate_question} and \Cref{appendix:data_creation}.

\subsection{Dataset Statistics}

\begin{figure*}[t]
    \centering
    \includegraphics[width=\linewidth]{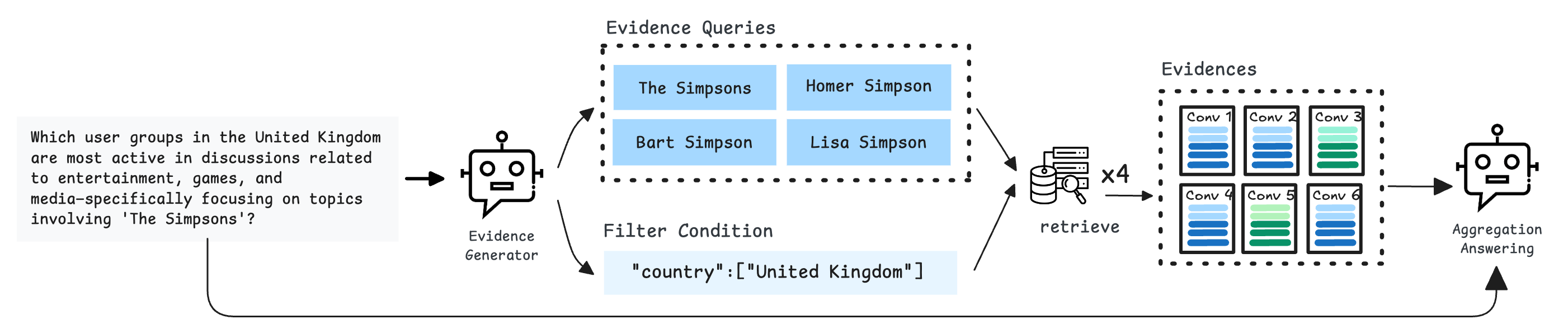}
    \caption{Overview of the PROBE retrieval approach.}
    \label{fig:probe-overview}
\end{figure*}

\begin{figure}[!t]
    \centering
    \begin{subfigure}[b]{0.45\textwidth}
        \includegraphics[trim={7px 5px 7px 5px},clip,width=1.0\linewidth]{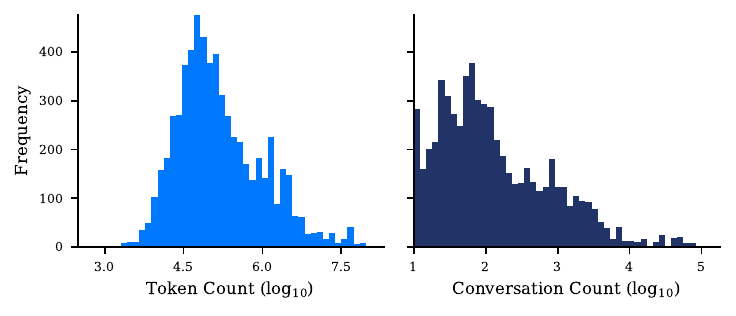}
    \end{subfigure}
    \caption{Distribution of total tokens and conversations in the supporting context.}
    \label{fig:context_token_count}
\end{figure}

The final WildChat-AQA benchmark contains 182,330 user-chatbot conversations and 6,027 aggregative questions. These conversations cover 28 high-level topics, 455 fine-grained subtopics, and 14,482 keyword categories. 
\Cref{tab:data_question_stat} in \Cref{sec:appendix_data_profile} provides detailed statistics of the questions organized by different attribute conditions and target attributes. Unlike typical question-answering tasks, which derive answers from one or a few documents, WildChat-AQA requires models to reason over contexts whose total token counts range widely from $\mathbf{10^1}$ to $\mathbf{10^8}$ tokens. \Cref{fig:context_token_count} illustrates the distribution of context token counts. Full data statistics are provided in \Cref{sec:appendix_data_profile}.


\subsection{Evaluation Protocol}
We frame the evaluation of aggregative question answering as a ranking problem. During training, the model or system is provided access to the entire WildChat-AQA dataset. At test time, the model is given an aggregative question along with 10 candidate answers. Its task is to rank these candidates according to their relevance to the question. We use standard ranking metrics \textbf{NDCG@1}, \textbf{NDCG@3}, \textbf{NDCG@5}, and \textbf{NDCG@10} to measure performance.


\subsection{Human Evaluation}
To evaluate the quality of our inferred attributes, we conduct a human evaluation measuring both inter-annotator agreement (human-human) and human-model agreement using Cohen's $\kappa$. Specifically, we randomly sample 100 examples each for level-1 (topic) and level-2 (subtopic) taxonomy labeling. Due to the multi-label nature of these tasks, we compute per-label agreement by treating each possible category as an independent binary labeling task. For subtopic evaluation, we additionally report macro-average agreement scores aggregated across all topics to provide a comprehensive view of annotation reliability. 

\begin{table}[!t]
  \centering
    \begin{tabular}{@{}lcc@{}}
      \toprule
      Name  & Human–Human $\kappa$ & Human-Model $\kappa$ \\
      \midrule
      Topic     & 0.581   & 0.617 \\
      \midrule
      Subtopic  &  0.576   & 0.609    \\
      \bottomrule
    \end{tabular}%
\caption{Average Cohen's $\kappa$ indicating agreement between human annotators (human-human) and between human annotations and model predictions (human-model).}
\label{tab:human_eval}
\end{table}

We found that Cohen's $\kappa$ for both topics and subtopics indicates moderate to substantial agreement~\citep{doi:10.1177/001316446002000104}, demonstrating a high degree of reliability between human annotations and model predictions.

\section{Probing Retrieval Of Broad Evidence}
Traditional retrieval methods, including those used in retrieval-augmented generation (RAG) \citep{NEURIPS2020_6b493230}, typically aim to identify a small set of highly specific, relevant documents. However, for Aggregative Question Answering, it is essential to identify a broader range of documents that collectively support reasoning about high-level aggregational insights. To address this unique requirement, we introduce a customized retrieval method, \textbf{P}robing \textbf{R}etrieval \textbf{O}f \textbf{B}road \textbf{E}vidence (\textbf{PROBE}). PROBE operates in two main steps:

\paragraph{Broad Query Generation}  Given a question $\mathbf{Q}$, we first prompt an LLM to generate a comprehensive set of short, diverse queries that may help retrieve a broad range of relevant documents. Specifically, the LLM generates a set of $n$ queries ${q_1, q_2, \dots, q_n}$ related to the question. Additionally, the model generates strict filtering conditions $\mathbf{F} = {f_1, f_2, \dots, f_m}$ to exclude documents clearly unrelated to the question. Formally, this process is defined as:
\begin{equation*}
   \mathbf{F}, \{q_1, q_2, \cdots, q_n \} = \text{LLM} (\mathbf{p}, \mathbf{Q}),
\end{equation*} where $\mathbf{p}$ represents the prompt. 

\paragraph{Evidence Aggregation and Generation}  Next, each generated query $q_i$ along with the filtering conditions $\mathbf{F}$ is used individually to retrieve relevant documents. This results in $n$ separate retrieval runs. We then aggregate these results by merging the retrieved document lists according to their retrieval relevance scores. If a document appears multiple times across different queries, we use max pooling to assign it the highest relevance score it received from any query. Finally, we select the top $k$ documents from this aggregated list as evidence.

The resulting set of retrieved documents serves as supporting evidence for the model to perform aggregational reasoning and answer the question. An overview of the full PROBE retrieval pipeline is in \Cref{fig:probe-overview}.



\section{Experiments}

\subsection{Models}
We experiment with widely-used models spanning various sizes: Gemma 3-4B~\citep{gemmateam2025gemma3technicalreport}, Qwen3-8B, Qwen3-32B~\citep{yang2025qwen3technicalreport}, and GPT-4.1-mini~\citep{OpenAIGPT4.1}. We also evaluate reasoning models including Qwen3-8B-think, Qwen3-32B-think, and o4-mini~\citep{OpenAI_o3_o4mini_2025}.

\subsection{Experimental Setups}
 We explore several experimental setups to investigate how effectively models leverage conversational data to answer aggregative questions:   

\paragraph{Textual Similarity} We use textual similarity score including BM25 and embedding-based consine similarity using \texttt{text-embedding-3-large} embeddings (denoted as Cosine Sim) to rank the response without context information. 

\paragraph{Model With No Context} The model directly answers questions without external inputs, relying solely on internal knowledge. This approach establishes baseline performance using only pre-existing knowledge. Due to resource constraints, we only evaluate this baseline using o4-mini, which is one of the strongest reasoning models.


\paragraph{Retrieval Augmented Generation (RAG)}  We use standard retrieval-augmented generation using OpenAI's \texttt{text-embedding-3-large} embeddings to retrieve relevant conversations as context. 

\paragraph{Finetuning}  We finetune pretrained models on the entire WildChat-AQA raw conversations and summaries to test whether fine-tuning on context can bring improvement to the QA tasks. 

\paragraph{PROBE}  For our retrieval method, PROBE,, query generation uses GPT-4.1-mini, and the retrieval relies on embeddings from OpenAI's \texttt{text-embedding-3-large} model. 

\subsection{Raw vs. Summarized Document}
Raw conversations are detailed but noisy (average 1,143.4 tokens each), whereas summarized conversations are more concise (average 21.5 tokens). Therefore, we experiment with both raw and summarized conversation inputs to investigate their effectiveness for aggregative question answering. Implementation details for experiments are provided in \Cref{appendix:implementation_details}.

\subsection{Main Results}

\begin{table*}[!t]
\centering
\resizebox{0.92\textwidth}{!}{
\begin{tabular}{@{}l|l|lccccc@{}}
\toprule
Model Name                                      &  Context                  &  Type & NDCG@1                        & NDCG@3                        & NDCG@5                        & NDCG@10                & \makecell{\# Input Token \\(Million)} \\ \midrule
Random                                          &  -                      &   -                              & 0.2501	                    & 0.3516                        & 0.4368                        & 0.6211                 &  -             \\ \midrule
BM25                                            &  -                      &   -                              & 0.2320                        & 0.3529                        & 0.4385                        & 0.6208                 &  -              \\ \midrule
Cosine Sim              &  -                      &   -                              & 0.2761                        & 0.3795                        & 0.4638                        & 0.6382                 &  -              \\ \midrule
o4-mini                                         &  -                      &   -                              & 0.3063	                    & 0.4017                        & 0.4805                        & 0.6488                 &  0.87          \\ \midrule
\multirow{2}{*}{Qwen3 8B}                       & \multirow{2}{*}{Finetune}  & Raw                              & 0.2694                        & 0.3739                        & 0.4589                        & 0.6346                 &  1.74          \\
                                                &                            & Summary                          & \underline{0.2984}	        & \underline{0.3966}	        & \underline{0.4807}	        & \underline{0.6480}     &  1.74          \\ \midrule 
\multirow{4}{*}{Gemma3 4B}                      & \multirow{2}{*}{RAG}       & Raw                              & 0.3291                        & 0.4356                        & 0.5159                        & 0.6688                 &  73.48         \\
                                                &                            & Summary                          & 0.3740                        & 0.4895                        & 0.5627                        & 0.6991                 &  174.62        \\
                                                & \multirow{2}{*}{PROBE}     & Raw                              & 0.4766                        & 0.5891                        & 0.6478                        & 0.7620                 &  38.44         \\
                                                &                            & Summary                          & \underline{0.5430}            & \underline{0.6513}            & \underline{0.6994}            & \underline{0.7969}     &  17.35         \\ \midrule
\multirow{4}{*}{\makecell{Qwen3 8B \\Think}}    & \multirow{2}{*}{RAG}       & Raw                              & 0.4168                        & 0.5090                        & 0.5779                        & 0.7123                 &  362.16        \\
                                                &                            & Summary                          & 0.5273                        & 0.6110                        & 0.6646                        & 0.7717                 &  176.88        \\
                                                & \multirow{2}{*}{PROBE}     & Raw                              & 0.6545                        & 0.7305                        & 0.7728                        & 0.8483                 &  315.52        \\
                                                &                            & Summary                          & \underline{0.6944}            & \underline{0.7638}            & \underline{0.8005}            & \underline{0.8660}     &  123.04        \\ \midrule
\multirow{4}{*}{\makecell{Qwen3 32B \\Think}}   & \multirow{2}{*}{RAG}       & Raw                              & 0.4052                        & 0.5020                        & 0.5705                        & 0.7081                 &  182.90        \\
                                                &                            & Summary                          & 0.5496                        & 0.6321                        & 0.6847                        & 0.7850                 &  176.88        \\
                                                & \multirow{2}{*}{PROBE}     & Raw                              & 0.6525                        & 0.7347                        & 0.7759                        & 0.8501                 &  315.52        \\
                                                &                            & Summary                          & \underline{0.7056}            & \underline{0.7753}            & \underline{0.8114}            & \underline{0.8725}     &  123.04        \\ \midrule
\multirow{4}{*}{GPT-4.1 mini}                   & \multirow{2}{*}{RAG}       & Raw                              & 0.4494                        & 0.5387                        & 0.6035                        & 0.7299                 &  344.37        \\
                                                &                            & Summary                          & 0.5782                        & 0.6620                        & 0.7104                        & 0.8019                 &  154.31        \\
                                                & \multirow{2}{*}{PROBE}     & Raw                              & 0.6806                        & 0.7536                        & 0.7936                        & 0.8628                 &  298.69        \\
                                                &                            & Summary                          & \underline{0.7308}            & \underline{0.7942}            & \underline{0.8282}            & \underline{0.8843}     &  107.11        \\ \midrule
\multirow{4}{*}{o4-mini}                        & \multirow{2}{*}{RAG}       & Raw                              & 0.4730                        & 0.5510                        & 0.6116                        & 0.7383                 &  344.37        \\
                                                &                            & Summary                          & 0.6122                        & 0.6792	                    & 0.7242                        & 0.8140                 &  154.31        \\
                                                & \multirow{2}{*}{PROBE}     & Raw                              & 0.7117                        & 0.7747                        & 0.8086                        & 0.8745                 &  298.69        \\
                                                &                            & Summary                          & \textbf{\underline{0.7571}}   & \textbf{\underline{0.8095}}	& \textbf{\underline{0.8386}}	& \textbf{\underline{0.8930}}  &  107.11        \\ \bottomrule
\end{tabular}}

\caption{Experiment results of different models using various retrieval approaches and conversation formats (raw vs. summarized). \underline{Underlined} scores indicate the best results for each model, and \textbf{bold} scores indicate the best overall results.}\label{tab:expr_main}
\end{table*}

\Cref{tab:expr_main} presents performance results across different models, retrieval methods, and conversation formats.

\paragraph{Simple textual relevance is ineffective.} We experiment with simple BM25 and embedding-based textual similarity models. We find that textual relevance baselines performed no better than random selection. The embedding-based approach performs slightly better than BM25, improving NDCG@1, 3, 5, and 10 by 4.41, 2.66, 2.53, and 1.74, respectively.

\paragraph{Stronger models perform better.}
 Among tested models, o4-mini consistently achieves the highest performance, with a maximum NDCG@1 score of 0.7571. GPT-4.1-mini, while also strong, trails slightly behind. Among open-source models, Qwen3-32B-think achieves the highest performance. (0.7056 NDCG@1).
 

\paragraph{PROBE outperforms standard RAG.}
Compared to standard RAG, PROBE consistently shows large performance improvements. With raw data, PROBE improves NDCG@1 scores by 14.8, 23.7, 24.7, 23.1, and 23.8 points for Gemma3-4B, Qwen3-8B-think, Qwen3-32B-think, GPT-4.1-mini, and o4-mini, respectively. A similar trend is observed using summarized conversations.


\paragraph{Summaries outperform raw conversations.}
Models consistently perform better with summarized inputs, showing improved NDCG@1 scores of 4.5 to 14.4 points over raw conversations for standard RAG, and 4.0 to 6.6 points for PROBE. Summaries enable more efficient information retrieval and easier aggregation of insights.

\paragraph{Basic finetuning is not effective.}
 Direct finetuning on Qwen3-8B (raw or summarized conversations without explicit aggregative reasoning steps) does not substantially exceed random-chance performance. This suggests that standard finetuning alone may be insufficient to internalize aggregative information. We caution, however, against generalizing this finding to all finetuning strategies: more sophisticated approaches that explicitly incorporate aggregative reasoning traces during training could yield stronger results, making this an important avenue for future work.


\paragraph{Token consumption is high.}
Achieving good performance on this task requires models to consume a very large number of input tokens as shown in \Cref{tab:expr_main}. This highlights a significant computational challenge and motivates future research to improve efficiency.

\begin{table*}[h]
\centering
\small
\begin{tabular}{@{}lccccccc@{}}
\toprule
\textbf{Method} & \textbf{R@5} & \textbf{R@10} & \textbf{R@20} & \textbf{R@50} & \textbf{R@100} & \textbf{R@200} & \textbf{R@500} \\
\midrule
RAG-Dense      & 0.01          & 0.02	       & 0.04	      & 0.07	     & 0.10         &	0.14     & 0.21 \\
\midrule
PROBE-Dense   & 0.07          & 0.13         & 0.23         & 0.35         & 0.43         & 0.50         & 0.58         \\
\quad - filter only     & 0.05 \textcolor{gray}{ (-0.02)} & 0.09 \textcolor{gray}{ (-0.04)} & 0.16 \textcolor{gray}{ (-0.07)} & 0.24 \textcolor{gray}{ (-0.11)} & 0.29 \textcolor{gray}{ (-0.14)} & 0.33 \textcolor{gray}{ (-0.17)} & 0.40 \textcolor{gray}{ (-0.18)} \\
\quad - question \& filter   & 0.06 \textcolor{gray}{ (-0.01)} & 0.12 \textcolor{gray}{ (-0.01)} & 0.21 \textcolor{gray}{ (-0.02)} & 0.32 \textcolor{gray}{ (-0.03)} & 0.40 \textcolor{gray}{ (-0.03)} & 0.46 \textcolor{gray}{ (-0.04)} & 0.53 \textcolor{gray}{ (-0.05)} \\
\bottomrule
\end{tabular}
\caption{Recall@k of PROBE-Dense (Summary) with ablations removing generated queries or filters. Numbers in parentheses indicate performance decrease compared to the full PROBE approach.}
\label{tab:recall_abalation}
\end{table*}

\begin{table}[h]
\small
\resizebox{0.49\textwidth}{!}{
\begin{tabular}{@{}clccc@{}}
\toprule
\# Conversation       & Context & Recall     & NDCG@5  & \# Input Token (M)       \\
\midrule
\multirow{3}{*}{5}   & RAG      & 0.01       & 0.5373       & 0.33 \\
                     & PROBE    & 0.07       & 0.6991       & 0.33 \\
                     & Oracle   & {\ul 0.10} & {\ul 0.7925} & 0.33 \\\midrule 
\multirow{3}{*}{20}  & RAG      & 0.04       & 0.5897       & 0.80 \\
                     & PROBE    & 0.23       & 0.7624       & 0.78 \\
                     & Oracle   & {\ul 0.34} & {\ul 0.8540} & 0.77 \\\midrule
\multirow{3}{*}{50}  & RAG      & 0.07       & 0.6318       & 1.74 \\
                     & PROBE    & 0.35       & 0.7927       & 1.60 \\
                     & Oracle   & {\ul 0.54} & {\ul 0.8721} & 1.46 \\\midrule
\multirow{3}{*}{200} & RAG      & 0.14       & 0.6858       & 6.46 \\
                     & PROBE    & 0.50       & 0.8202       & 5.13 \\
                     & Oracle   & {\ul 0.75} & {\ul 0.8942} & 3.63 \\\midrule
\multirow{3}{*}{500} & RAG      & 0.20       & 0.7141       & 15.4 \\
                     & PROBE    & 0.58       & 0.8263       & 11.3 \\
                     & Oracle   & {\ul 0.84} & {\ul 0.9005} & 6.31 \\\bottomrule
\end{tabular}}

\caption{NDCG@5 scores, recall rates, and input lengths (in millions of tokens) using o4-mini with summarized conversations. \underline{Underlined} values indicate the best score for each number of conversations.}
\label{tab:recall_vs_ndcg_table}
\end{table}

\subsection{Ablation Studies}
We conduct ablation studies on a stratified 10\% subset of the benchmark, selected based on the condition and target types. 

\paragraph{Retrieval effectiveness is crucial.}
Retrieval quality substantially affects final performance. \Cref{tab:recall_vs_ndcg_table} reports the results of o4-mini under varying recall rates from different retrieval methods. Higher recall rates consistently yield better NDCG scores. 

\begin{figure}[!t]
     \centering
     \includegraphics[width=0.9\linewidth]{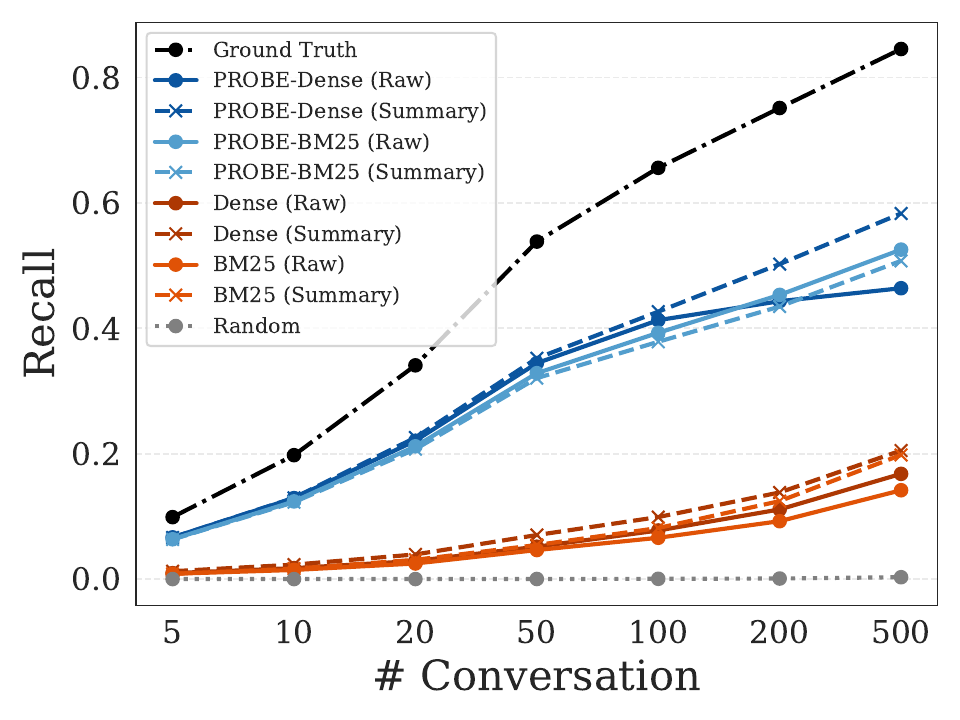}
     \caption{Recall of different retrieval approaches.}
     \label{fig:recall_ablation}
\end{figure}

\paragraph{Retrieval performance.} 
We compare various retrieval approaches, including vector-based embeddings, BM25, random, and ground-truth retrieval. \Cref{fig:recall_ablation} shows recall rates for different retrieval strategies. PROBE consistently provided substantial improvements over standard RAG, with the highest recall from PROBE-Dense (summarized). Removing either the generated query or filtering steps notably degrades PROBE's retrieval effectiveness as shown in \Cref{tab:recall_abalation}.
 
\paragraph{Existing models lack effective aggregational reasoning capabilities.}
To evaluate model capabilities under ideal conditions, we perform experiments using oracle documents as context. \Cref{tab:expr_oracle} shows that all models perform better when given summarized contexts rather than raw conversations, indicating challenges in aggregating information from longer, noisier texts.

\begin{table}[!t]
\resizebox{\linewidth}{!}{
\begin{tabular}{@{}llcccc@{}}
\toprule
Model Name                      & Ctx Type & NDCG@1 & NDCG@3 & NDCG@5 & NDCG@10  \\ \midrule
\multirow{2}{*}{Gemma3 4B}      & Raw      & 0.4815	& 0.6057 & 0.6601 & 0.7703    \\
                                & Summary  & 0.5699	& 0.6787 & 0.7235 & 0.8102    \\ \midrule
\multirow{2}{*}{Qwen3 8B Think} & Raw      & 0.7359	& 0.7991 & 0.8360 & 0.8894    \\
                                & Summary  & 0.7757 & 0.8268 & 0.8510 & 0.9003    \\ \midrule
\multirow{2}{*}{Qwen3 32B Think}& Raw      & 0.7225 & 0.8044 & 0.8355 & 0.8897    \\
                                & Summary  & 0.8134 & 0.8605 & 0.8817 & 0.9199    \\ \midrule
\multirow{2}{*}{GPT-4.1-mini}   & Raw      & 0.7849	& 0.8388 & 0.8667 & 0.9121    \\
                                & Summary  & 0.8130 & 0.8602 & 0.8816 & 0.9216    \\ \midrule
\multirow{2}{*}{o4-mini}        & Raw      & 0.8003 & 0.8456 & 0.8719 & 0.9185    \\
                                & Summary  & \textbf{0.8478} & \textbf{0.8793} & \textbf{0.9005} & \textbf{0.9347}   \\
\bottomrule
\end{tabular}}
\caption{Results of aggregative question answering with oracle (ground-truth) documents as context.}\label{tab:expr_oracle}
\end{table}


\begin{figure}[!t]
    \centering
    \includegraphics[width=0.98\linewidth]{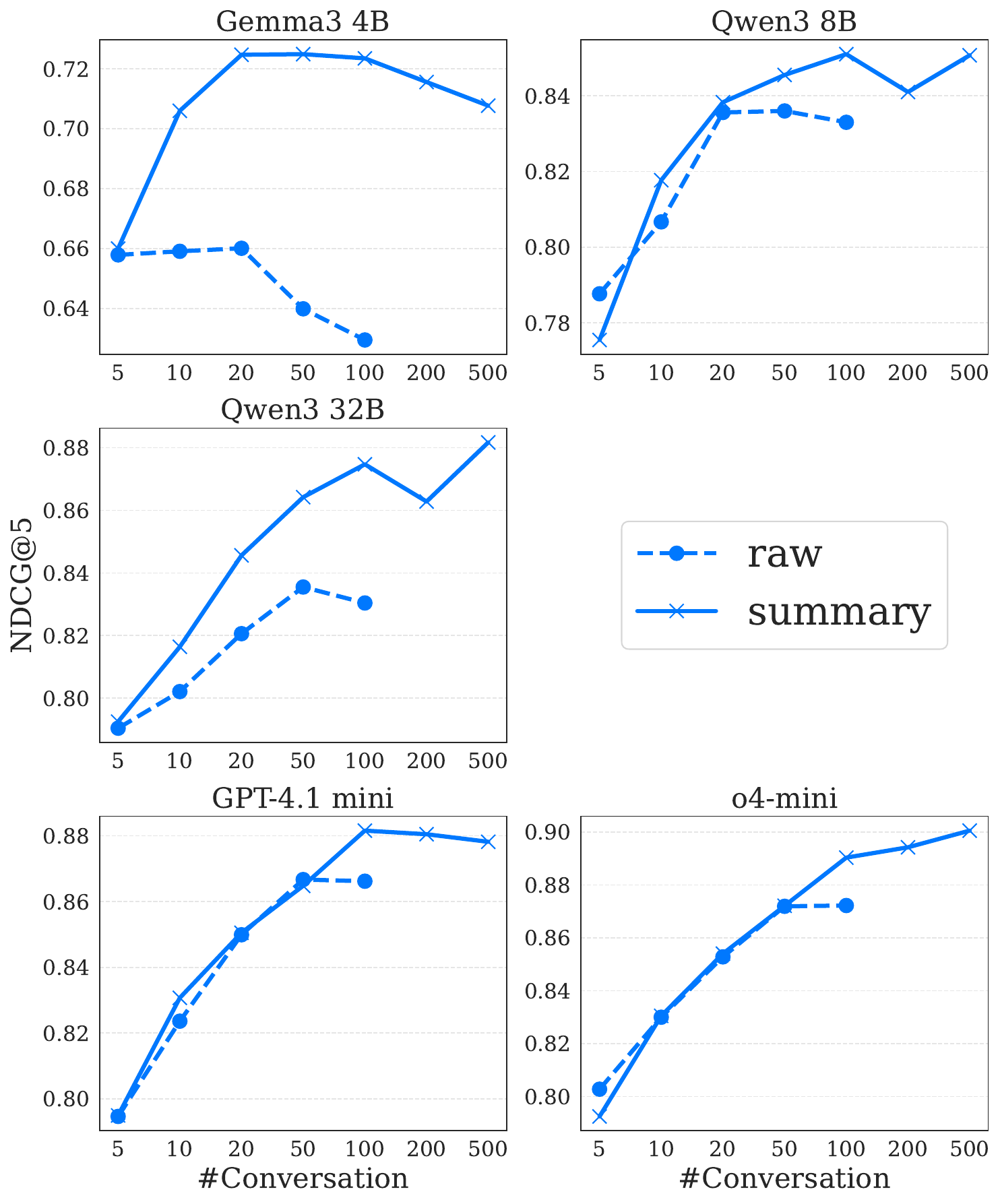}
\caption{NDCG@5 scores for different models given varying numbers of oracle (ground-truth) documents, comparing raw and summarized conversations.}    \label{fig:oracle_experiment}

\end{figure}

We further analyze how performance varies with the number of provided conversations (\Cref{fig:oracle_experiment}). Weaker models such as Gemma3 and Qwen3 show a substantial performance gap between raw and summarized contexts, even when given the same number of conversations, highlighting their limited ability to implicitly extract relevant information. Stronger models like GPT-4.1-mini and o4-mini show a smaller initial gap, but this gap widens notably when the context is extended to 100 documents, demonstrating that even advanced models struggle with aggregating and reasoning effectively over extensive raw contexts.

\begin{figure}[!t]
    \centering
    \includegraphics[width=0.98\linewidth]{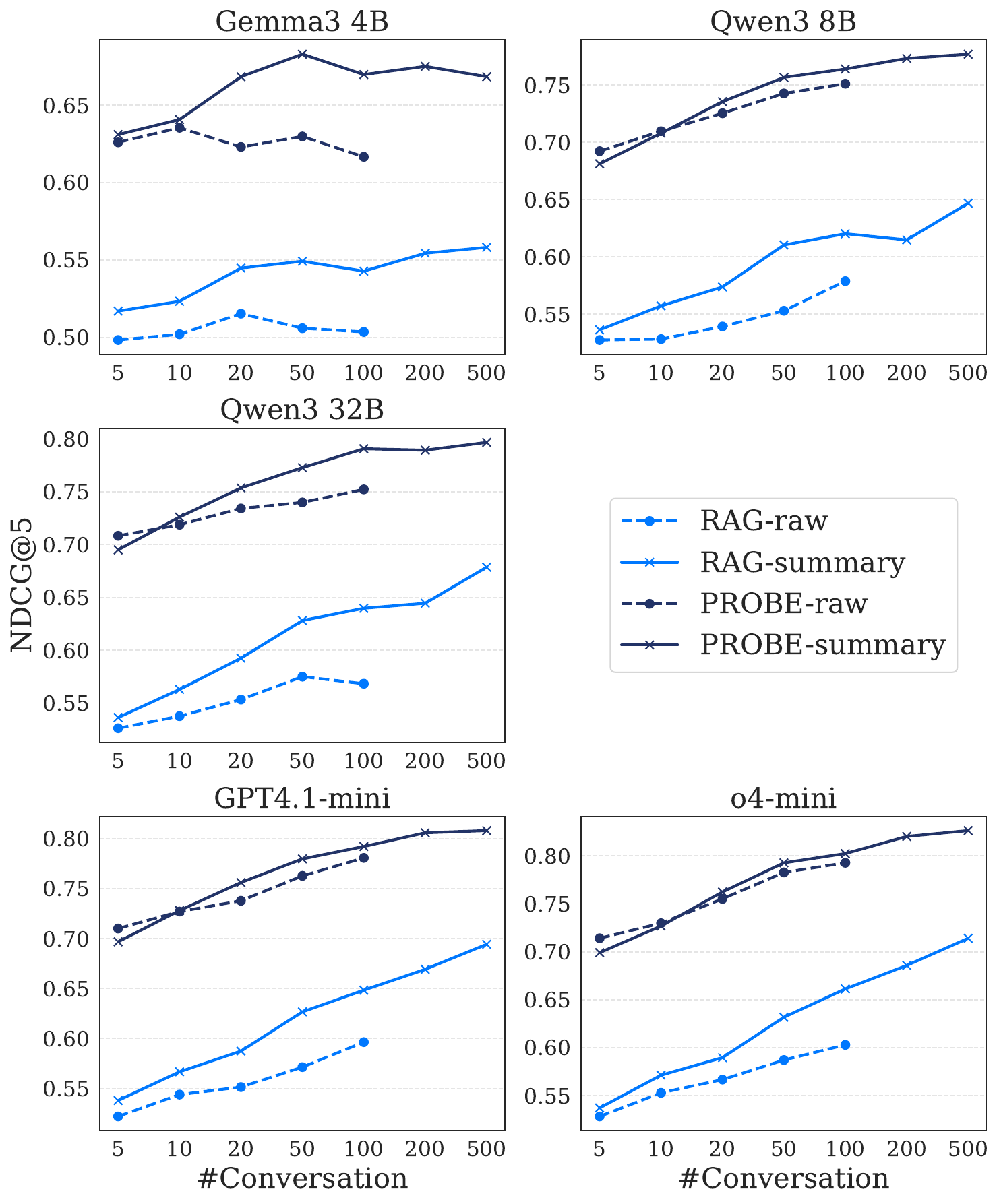}
\caption{Comparison of NDCG@5 scores for different models with varying numbers of retrieved documents.}
\label{fig:retrieved_ndcg_comp}
\end{figure}

\paragraph{Performance improves with more context.}
Unlike standard RAG tasks, Aggregative Question Answering fundamentally relies on a broader set of documents. As more documents are provided, models improve significantly in answering aggregative questions (\Cref{fig:retrieved_ndcg_comp}). This finding validates that aggregative question answering requires extensive context and global dataset knowledge.

\Cref{fig:oracle_experiment,fig:retrieved_ndcg_comp} show that under all experiment settings, performance improves as more documents are provided, demonstrating the necessity of incorporating global information from the dataset.

\paragraph{Aggregative question answering is reasoning-intensive.}
We evaluate Qwen3-32B with the ``think'' mode on to measure the effect of explicit reasoning. The results (see \Cref{tab:ablation_thinking}) consistently show reasoning led to significant performance improvements across all experimental setups, indicating aggregative question answering demands substantial reasoning abilities.

\begin{table}[!t]
\centering
\resizebox{\linewidth}{!}{
\begin{tabular}{@{}lcccc@{}}
\toprule
\textbf{Method}       & \textbf{NDCG@1} & \textbf{NDCG@3} & \textbf{NDCG@5} & \textbf{NDCG@10} \\
\midrule
Oracle                & 0.72            & 0.79            & 0.82            & 0.88            \\
\quad + thinking      & 0.81  \textcolor{gray}{(+0.09)} & 0.86  \textcolor{gray}{(+0.07)} & 0.88  \textcolor{gray}{(+0.06)} & 0.92  \textcolor{gray}{(+0.04)} \\
\addlinespace\midrule
RAG (Summary)         & 0.48            & 0.56            & 0.62            & 0.74            \\
\quad + thinking      & 0.54  \textcolor{gray}{(+0.07)} & 0.62  \textcolor{gray}{(+0.07)} & 0.68  \textcolor{gray}{(+0.06)} & 0.78  \textcolor{gray}{(+0.04)} \\
\addlinespace \midrule
PROBE (Summary)       & 0.64            & 0.71            & 0.75            & 0.84            \\
\quad + thinking      & 0.68  \textcolor{gray}{(+0.04)} & 0.76  \textcolor{gray}{(+0.05)} & 0.80  \textcolor{gray}{(+0.05)} & 0.86  \textcolor{gray}{(+0.02)} \\
\bottomrule
\end{tabular}}
\caption{NDCG scores of Qwen3 32B with and without reasoning (``think'' mode). Improvements from reasoning are indicated in parentheses.}\label{tab:ablation_thinking}
\end{table}

\section{Future Research Directions}

\paragraph{Reasoning Over Very Long Context}
In this work, we experiment with several reasoning-capable models and observe that current models typically have limited context windows, and performance degrades sharply as the length of the input context increases. Developing efficient and accurate methods for reasoning over very long textual contexts remains an important open problem.

\paragraph{Cost-Efficient Aggregative Question Answering}
Current effective solutions for Aggregative Question Answering require processing extremely large amounts of text, resulting in substantial computational costs. Future research could explore hierarchical indexing, retrieval strategies, and long-term memory mechanisms to reduce token consumption and improve computational efficiency.


 
\paragraph{Streaming Aggregative Question Answering}  In real-world scenarios, chatbot conversations often arrive in continuous streams rather than static collections. Future research could explore methods to dynamically update aggregational insights as new interactions occur in real time. Ideally, conversational agents would continuously integrate information from ongoing interactions, similar to how humans update their understanding based on new experiences, to maintain up-to-date and adaptive aggregational knowledge.


\section{Conclusion}
In this paper, we introduce Aggregative Question Answering, a new task aimed at extracting collective insights from large-scale conversational data generated by interactions between users and LLM-powered chatbots. To facilitate research in this area, we construct the WildChat-AQA benchmark, comprising 6,027 aggregational questions derived from 182,330 real-world chatbot conversations. Our experiments demonstrate that existing state-of-the-art methods, including fine-tuning, Retrieval-Augmented Generation (RAG), and even an improved RAG approach specifically adapted for this task struggle significantly, either failing to reason effectively at the necessary global scale or incurring prohibitively high computational costs. Looking ahead, we believe that addressing these challenges would enable future models to better derive meaningful user and societal insights from large-scale conversational data.


\section*{Limitations}
\paragraph{Potential Errors in Model-derived Annotations}
Although we employ powerful large LLMs and pipelines such as GPT-4o and TnT-LLM to infer attributes and assign taxonomy labels, errors and inconsistencies may occur due to model hallucinations or instruction misalignment. Specifically, hallucinations might affect both the inferred topics (summaries used to construct taxonomies) and the extracted keywords, potentially introducing noise or inaccuracies into the benchmark. To quantify these potential errors, we conduct human evaluations measuring the agreement between human annotations and LLM annotations for both topic extraction (Table 2) and keyword extraction. Although these evaluations confirm moderate to high accuracy, we acknowledge that some errors remain inevitable. Additionally, real-world conversational data are inherently noisy, ambiguous, and challenging to categorize neatly, making completely error-free annotations unattainable. We encourage future users of our dataset to remain aware of these limitations when interpreting experimental results.

\paragraph{Artificiality of Generated Questions}
Aggregative questions in WildChat-AQA were generated by prompting GPT-4.1 to translate structured database queries into natural-language questions. While effective and typically resulting in simple and straightforward queries, this method may introduce stylistic, syntactic, and semantic artifacts. Models trained on our data can potentially overfit to the stylistic patterns of LLM-generated questions, which could limit the validity of the introduced benchmark. Consequently, strong performance on WildChat-AQA may not directly generalize to success on genuinely human-authored aggregative questions, which tend to be linguistically richer and more diverse. We thus consider strong performance on our benchmark as a necessary but not sufficient condition for aggregative question-answering capabilities in real-world scenarios.



\section*{Ethical Considerations}
Aggregative Question Answering opens promising avenues for real-world analytics but also raises potential ethical and societal concerns, particularly when insights relate to sensitive topics such as elections, public opinion, or public health---areas that could potentially be susceptible to manipulation. To reduce the risks of reinforcing stereotypes or enabling sensitive demographic profiling, we avoided constructing questions targeting protected attributes. Moreover, all experiments conducted in this work rely exclusively on the publicly available and anonymized WildChat dataset, which is explicitly intended for open research purposes (licensed under ODC-BY). By introducing WildChat-AQA as an open benchmark, we aim to empower transparent academic research that responsibly explores both the capabilities and risks associated with aggregational analytics. Our goal is to encourage the open research community to evaluate these powerful systems, rather than relying solely on proprietary analyses conducted behind closed doors.

\section*{Acknowledgements}
Yuntian Deng acknowledges support from an NSERC Discovery Grant (RGPIN-2024-05178), a Starter Grant from the University of Waterloo, and research funding provided by Manulife through the Waterloo Data and Artificial Intelligence Institute. We also thank OpenAI's Research Access Program for providing API credits. Wentao Zhang is supported in part by these sources and by the Dr. Derick Wood Graduate Scholarship, generously funded by Ms. Mary Chen.


\bibliography{custom}
\appendix

\clearpage

\section{Related Works}

\paragraph{Question Answering} Question answering typically involves a diverse range of perspectives. Datasets such as TriviaQA \cite{joshi-etal-2017-triviaqa}, RACE \cite{lai-etal-2017-race}, HotPotQA \cite{yang2018hotpotqa}, Natural Questions \cite{kwiatkowski-etal-2019-natural}, MuSiQue \cite{trivedi-etal-2022-musique}, 2Wiki \cite{ho-etal-2020-constructing}, PopQA \cite{mallen2023trustlanguagemodelsinvestigating}, and MultiHop-RAG \cite{tang2024multihopragbenchmarkingretrievalaugmentedgeneration} focus on \textbf{local information}, where answers can be derived from one or several documents.
In contrast, other benchmarks such as MMLU \cite{hendrycks2021measuring}, MATH \cite{hendrycksmath2021}, GSM8K \cite{cobbe2021gsm8k}, and Big-Bench \cite{srivastava2023beyond} emphasize science, technology, engineering, mathematics, and logical reasoning. These primarily evaluate models' world knowledge and reasoning capabilities but lack a benchmark for understanding large-scale datasets and deriving high-level insights.
Recent works such as GraphRAG \cite{edge2025localglobalgraphrag} address the long-context challenge by extracting entities and relationships from extended text data and constructing graph structures to answer questions.

\paragraph{Long Context Retrieval Augmented Generation}  \cite{NEURIPS2020_6b493230} has emerged as a prominent approach for enhancing the performance of large language models (LLMs) on knowledge-intensive tasks while also mitigating hallucinations. Recently, advances in computational capabilities have spurred interest in extending RAG to support very long contexts. Several studies—such as those by \citet{jiang2024longragenhancingretrievalaugmentedgeneration}, \citet{zhao-etal-2024-longrag}, and \citet{jin2025longcontext}—have proposed methods to improve the effectiveness of LLMs in long-context settings. In parallel, \citet{lee2024longcontextlanguagemodelssubsume} introduced LOFT, a new benchmark designed to evaluate LLMs on a broad range of tasks addressable by either RAG or long-context modeling.

\paragraph{Summarization} Summarization has been a long-standing challenge in natural language processing. Early benchmark datasets, such as CNN/Daily Mail \cite{see-etal-2017-get} and XSum \cite{Narayan2018DontGM}, primarily targeted single-document summarization. Subsequent efforts, including MultiNews \cite{fabbri-etal-2019-multi} and MS$^2$ \cite{deyoung-etal-2021-ms}, extended this task to the multi-document setting. Another line of related work focuses on query-based summarization, for which QMSum \cite{zhong-etal-2021-qmsum} and DUC 2005 \cite{10.5555/1654679.1654689} are two widely used datasets.

\paragraph{Text to SQL} Text-to-SQL is a widely studied approach for tackling aggregative question answering. In this paradigm, the model is required to generate a structured database query based on a natural language question. Several established benchmarks have been proposed to evaluate this task, including WikiSQL \cite{zhongSeq2SQL2017}, Spider \cite{lei2024spider}, BIRD \cite{li2024can}, and WikiTableQA \cite{pasupat-liang-2015-compositional}. Additionally, LOFT \cite{lee2024longcontextlanguagemodelssubsume} includes a sub-task specifically designed to assess how effectively large language models can emulate database-style querying.

\section{Data Statistics}
\label{sec:appendix_data_profile}

\subsection{Statistics of Generated Question by Condition and Targets}
\begin{table}[ht]
\centering
\begin{tabularx}{\columnwidth}{>{\raggedright\arraybackslash}X l r}
\toprule
Condition & Target & Count \\
\midrule
 \multicolumn{3}{c}{0 Condition}   \\
\midrule
none & topic & 1 \\ 
none & loc & 1 \\ 
none & lang & 1 \\ 
\midrule
 \multicolumn{3}{c}{1 Condition}   \\
\midrule
user & keywords & 370 \\ 
user & time & 100 \\ 
keywords & user & 96 \\ 
user & lang & 60 \\ 
user & topic & 54 \\ 
time & user & 39 \\ 
topic & subtopic & 26 \\ 
loc & topic & 20 \\ 
loc & keywords & 17 \\ 
lang & topic & 9 \\ 
time & topic & 6 \\ 
time & keywords & 6 \\ 
topic & loc & 6 \\ 
topic & user & 6 \\ 
topic & lang & 4 \\ 
topic & keywords & 4 \\ 
time & lang & 4 \\ 
lang & keywords & 1 \\ 
\midrule
 \multicolumn{3}{c}{2 Conditions}   \\
\midrule
user, topic & subtopic & 199 \\ 
user, topic & keywords & 185 \\ 
user, user & subtopic & 141 \\ 
user, topic & time & 114 \\ 
topic, lang & subtopic & 100 \\ 
time, topic & user & 98 \\ 
time, topic & subtopic & 98 \\ 
topic, lang & user & 98 \\ 
topic, loc & time & 97 \\ 
topic, keywords & user & 97 \\ 
\bottomrule
\end{tabularx}
\caption{Question Type Statistics}
\label{tab:data_question_stat}
\end{table}

\begin{table}[!t]
\centering
\begin{tabularx}{\columnwidth}{>{\raggedright\arraybackslash}X l r}
\toprule
Condition & Target & Count \\
\midrule
topic, loc & subtopic & 96 \\ 
topic, keywords & time & 96 \\ 
time, user & keywords & 94 \\ 
topic, subtopic & user & 93 \\ 
subtopic, subtopic & user & 93 \\ 
topic, loc & keywords & 82 \\ 
topic, lang & time & 74 \\ 
time, topic & loc & 60 \\
topic, subtopic & keywords & 55 \\ 
topic, topic & user & 55 \\ 
time, user & topic & 53 \\ 
user, user & topic & 53 \\ 
time, topic & keywords & 49 \\ 
topic, subtopic & loc & 39 \\ 
time, loc & topic & 34 \\ 
time, lang & topic & 31 \\ 
topic, lang & keywords & 27 \\ 
time, topic & lang & 15 \\ 
topic, subtopic & lang & 13 \\ 
topic, loc & user & 10 \\ 
\midrule
 \multicolumn{3}{c}{3 Conditions}   \\
\midrule
loc, topic, subtopic & user & 287 \\ 
lang, topic, subtopic & user & 284 \\ 
user, topic, subtopic & keywords & 276 \\ 
time, loc, topic & user & 199 \\ 
time, topic, subtopic & keywords & 175 \\ 
user, user, user & subtopic & 132 \\ 
user, topic, keywords & time & 114 \\ 
time, topic, keywords & user & 100 \\ 
time, loc, topic & subtopic & 100 \\ 
time, user, topic & subtopic & 100 \\ 
loc, topic, keywords & user & 99 \\ 
user, topic, subtopic & time & 98 \\ 
user, topic, keywords & subtopic & 98 \\ 
loc, topic, keywords & time & 98 \\ 
lang, topic, keywords & time & 98 \\ 
time, topic, subtopic & user & 97 \\ 
lang, topic, keywords & user & 96 \\ 
topic, subtopic, keywords & user & 94 \\ 
loc, topic, subtopic & keywords & 93 \\ 
lang, topic, subtopic & keywords & 82 \\ 
time, topic, subtopic & loc & 76 \\ 
user, user, user & topic & 51 \\ 

\bottomrule
\end{tabularx}
\end{table}

\subsection{Language Distribution}
We provide a statistics of all language involved in the conversations in \Cref{tab:all_language_statistics}.
\begin{table*}[ht]
\centering

\begin{tabular}{lrlrlrlr}
\toprule
\textbf{Language} & \textbf{Count} & \textbf{Language} & \textbf{Count} & \textbf{Language} & \textbf{Count} & \textbf{Language} & \textbf{Count} \\
\midrule
English & 124,646 & Spanish & 4,193 & Italian & 744 & Polish & 527 \\
Russian & 22,877 & Portuguese & 3,532 & Korean & 605 & Vietnamese & 463 \\
Chinese & 6,434 & Turkish & 1,408 & Indonesian & 566 & Ukrainian & 406 \\
French & 4,782 & Latin & 1,239 & Dutch & 549 & Other & 1,824 \\
German & 4,487 & Arabic & 863 & Tagalog & 537 & & \\
\bottomrule
\end{tabular}
\caption{Language Statistics in Conversations}
\label{tab:all_language_statistics}
\end{table*}

\subsection{Keywords Cloud}
\begin{figure*}[h]
    \centering
    \includegraphics[width=\linewidth]{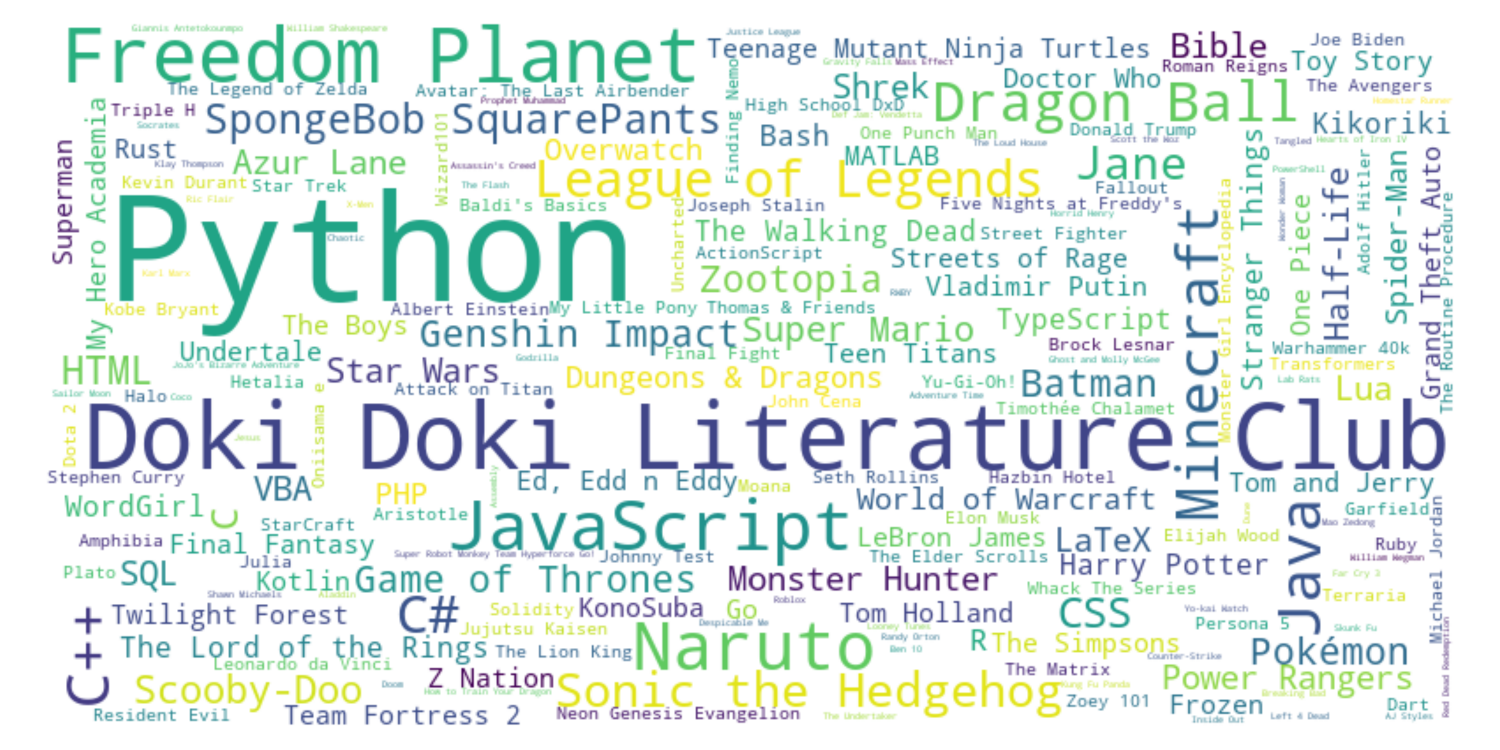}
    \caption{Word Cloud of All Keywords}
    \label{fig:keyword_cloud}
\end{figure*}

To illustrate the result of keywords categorization, we build a keywords cloud in \Cref{fig:keyword_cloud}.
\subsection{Topic and Subtopic Overview}
\clearpage

\small 
\onecolumn
\begin{longtable}{p{6.5cm}p{6.5cm}r}
\caption{Topic Taxonomy in WildChat-AQA}\label{tab:topic_taxonomy_full}\\
\toprule
\textbf{Parent Topic} & \textbf{Sub-topic} & \textbf{Count}\\
\midrule
\endfirsthead
\multicolumn{3}{c}{\textit{Topic Taxonomy in WildChat-AQA (continued)}}\\
\midrule
\textbf{Parent Topic} & \textbf{Sub-topic} & \textbf{Count}\\
\midrule
\endhead
\midrule
\multicolumn{3}{r}{\textit{Continued on next page}}\\
\endfoot
\midrule
\endlastfoot
\multirow{13}{*}{Creative Writing and Fiction} & Dialogue \& Scripted Scenes & 25421\\
 & Fanfiction \& Universe Crossovers & 20323\\
 & Extended Narrative Prose & 19771\\
 & Humorous \& Satirical Narratives & 11901\\
 & Erotic \& Sensual Narratives & 8304\\
 & World-Building \& Adventure Narratives & 6470\\
 & Creative Naming \& Prompt Generation & 4388\\
 & Sports \& Competition Narratives & 3370\\
 & Transformation \& Identity Narratives & 3283\\
 & Character Profiles \& Descriptions & 2025\\
 & Fictional News \& Media Formats & 1912\\
 & Poetic \& Lyric Composition & 1608\\
 & Interactive \& Roleplaying Narratives & 827\\
\midrule
\multirow{16}{*}{Law, Regulation and Criminal Justice} & Violent Crimes & 630\\
 & Regulatory Compliance and Licensing & 454\\
 & Civil Litigation and Consumer Protection & 284\\
 & Employment and Labor Law & 198\\
 & Sexual Crimes & 183\\
 & Intellectual Property and Copyright & 163\\
 & Financial, Fraud, and Cyber Offenses & 142\\
 & Robbery, Theft, and Property Offenses & 130\\
 & Judicial Process and Court Administration & 117\\
 & Constitutional Rights and Civil Liberties & 81\\
 & Terrorism, War Crimes, Treason, and Political Violence & 68\\
 & Corruption and Abuse of Power & 64\\
 & Public Order Offenses & 54\\
 & Immigration and Border Control & 51\\
 & Drug-Related Offenses & 50\\
 & Family and Marital Law & 48\\
\midrule
\multirow{13}{*}{Entertainment, Games, and Media} & Fanfiction \& Crossovers & 25629\\
 & Original Fiction \& Scripts & 4834\\
 & NSFW \& Explicit Scenes & 3717\\
 & Live-Action Film \& TV & 2963\\
 & Western Animation \& Comics & 2048\\
 & Gaming Story \& Lore & 1895\\
 & Celebrity \& Pop Culture & 1882\\
 & Gaming Mechanics \& Tech & 1660\\
 & Music \& Stage & 1651\\
 & Sports, eSports, \& Pro Wrestling & 1557\\
 & Anime \& Manga & 1552\\
 & Production \& Broadcasting & 1044\\
 & Tabletop \& TTRPG & 804\\
\midrule
\multirow{21}{*}{Software, Programming and Computer Science} & Programming & 17413\\
 & Web Development & 3603\\
 & AI and Machine Learning & 2787\\
 & Cybersecurity & 1930\\
 & Game Development, Design, and Modding & 1737\\
 & Databases and Queries & 1724\\
 & Operating Systems and Administration & 1414\\
 & Productivity and Desktop Software & 1215\\
 & Computer Networking & 1176\\
 & DevOps and Cloud & 1083\\
 & Data Analysis, Visualization and Business Intelligence & 1031\\
 & Mobile Development and Mobile Apps & 972\\
 & Computer Graphics & 740\\
 & Computer Science Theory & 612\\
 & Computer Hardware, Architecture, and Peripherals & 576\\
 & Software Architecture and Software System Design & 438\\
 & Testing and Quality Assurance & 350\\
 & Blockchain and Cryptocurrency & 336\\
 & Embedding Systems and IoT & 286\\
 & Human Computer Interaction & 184\\
 & Software Development Methodology and Project Management & 165\\
\midrule
\multirow{14}{*}{Science, Mathematics and Logical Reasoning} & Physics: Mechanics, Thermodynamics, and Fields & 1877\\
 & Basic Arithmetic and Numbers & 1376\\
 & Organismal Biology and Evolution & 1360\\
 & General Chemistry and Reactions & 1339\\
 & Cellular and Medical Sciences & 1239\\
 & Astronomy and Astrophysics & 1130\\
 & Earth Science and Environment & 1031\\
 & Statistics and Probability & 912\\
 & Algebra and Vectors & 833\\
 & Logic and Puzzles & 795\\
 & Geometry and Trigonometry & 724\\
 & Computational Science and Modeling & 610\\
 & Calculus and Higher Mathematics & 505\\
 & Materials, Engineering, and Technology & 363\\
\midrule
\multirow{20}{*}{Personal Advice and Support} & Navigating Romance and Dating & 464\\
 & Enhancing Personal Growth and Discipline & 286\\
 & Building Communication and Social Skills & 164\\
 & Offering Emotional Support and Love & 137\\
 & Navigating Sexual Intimacy, Consent, and Well-Being & 128\\
 & Supporting Mental Health and Well-Being & 111\\
 & Guiding Family, Parenting, and Caregiving & 99\\
 & Boosting Self-Confidence and Esteem & 81\\
 & Handling Career and Workplace Challenges & 73\\
 & Exploring Personal Values and Choices & 70\\
 & Seeking Apologies, Forgiveness, and Trust & 65\\
 & Addressing Financial Management and Housing & 47\\
 & Improving Physical Health and Body Image & 47\\
 & Managing Unwanted Contact and Boundaries & 38\\
 & Seeking Legal Guidance and Protective Measures & 34\\
 & Embracing Identity and Lifestyle Transitions & 32\\
 & Recovering from Breakups and Heartache & 32\\
 & Handling Emergencies, Threats, or Crises & 30\\
 & Overcoming Addictions and Harmful Habits & 19\\
 & Coping with Grief and Loss & 15\\
\midrule
\multirow{13}{*}{Business, Commerce and Finance} & Digital Marketing \& Social Media & 4010\\
 & Investments \& Financial Markets & 934\\
 & Business Operations \& Quality Management & 914\\
 & Accounting \& Financial Reporting & 891\\
 & Economic Trends \& Macro Outlook & 739\\
 & Corporate Governance \& Leadership & 492\\
 & Customer Service \& Complaints & 460\\
 & Legal \& Regulatory Compliance & 435\\
 & Supply Chain \& Logistics & 426\\
 & Wholesale \& B2B Distribution & 404\\
 & Banking \& Monetary Policies & 402\\
 & Careers \& Professional Development & 373\\
 & Entrepreneurship \& Startups & 356\\
\midrule
\multirow{26}[-40]{*}{History and Culture} & Modern and Contemporary History (19th Century–Present) & 1407\\
 & Conflicts and Wars & 1088\\
 & Medieval Europe & 716\\
 & Philosophy and Political Ideologies & 624\\
 & Art, Architecture, and Heritage & 616\\
 & Religion and Theology & 513\\
 & Traditions, Customs, and Rituals & 395\\
 & Popular Culture and Mass Media & 388\\
 & Pre-Modern East Asia & 386\\
 & Colonialism, Imperialism, and Independence & 343\\
 & Ancient Non-Classical Civilizations & 322\\
 & Classical Rome & 269\\
 & Diplomacy and Treaties & 251\\
 & Language and Literature & 240\\
 & Archaeology and Ancient Technologies & 217\\
 & Sports and Leisure & 197\\
 & Civil Rights and Social Justice & 192\\
 & Ancient Greece and Hellenic Culture & 174\\
 & Legal Systems and Codes & 172\\
 & Social Hierarchies and Slavery & 170\\
 & Myths and Folklore & 166\\
 & Gender and Women’s History & 166\\
 & Indigenous Peoples & 157\\
 & Science and Medicine & 154\\
 & Islamic and Middle Eastern Empires & 119\\
 & Exploration and Discoveries & 100\\
\midrule
\multirow{22}{*}{Lifestyle and Hobbies} & Exploring fashion and accessories & 204\\
 & Hair and Personal Grooming & 189\\
 & Beauty, makeup, and self-care & 110\\
 & Health, sports, and active living & 107\\
 & Minimalist living and conscious habits & 95\\
 & Personal expression, identity, and body positivity & 81\\
 & Creative crafts and DIY projects & 67\\
 & Outdoor Recreation and Camping & 61\\
 & Relationships, family, and social bonding & 59\\
 & Pets, animals, and responsible care & 46\\
 & Spirituality, meditation, and mindfulness & 45\\
 & Music, dance, and performing arts & 43\\
 & Games, collecting, and playful hobbies & 42\\
 & Social events, parties, and gatherings & 40\\
 & Costumes and cosplay & 37\\
 & Cooking, baking, and culinary hobbies & 31\\
 & Productivity and time management & 30\\
 & Travel, tourism, and new adventures & 24\\
 & Digital lifestyle and social media presence & 24\\
 & Seasonal festivities and holiday decorating & 12\\
 & Gardening and horticulture & 7\\
 & Home organization and interior comfort & 6\\
\midrule
\multirow{20}{*}{Academic Resource, Education and Learning} & Academic Research, Methods, and Presentation & 801\\
 & Curriculum and Course Development & 697\\
 & STEM and Technical Education & 428\\
 & Teaching Strategies and Pedagogical Tools & 423\\
 & Health and Medical Education & 326\\
 & Technology and AI Integration in Education & 296\\
 & Professional and Vocational Training & 248\\
 & Educational Policy and Leadership & 195\\
 & University Admissions and Scholarship Guidance & 157\\
 & Language Learning and Translation & 135\\
 & Memory, Study, and Exam Strategies & 118\\
 & Creative Arts and Literature in Education & 110\\
 & Early Childhood Education and Development & 104\\
 & Special Education and Inclusive Learning & 66\\
 & Socio-Emotional Learning and Wellbeing & 60\\
 & Environmental and Social Education & 43\\
 & Academic Ethics and Publication Guidelines & 34\\
 & Parental Engagement and Child Education & 34\\
 & Classroom Management and Student Engagement & 25\\
 & Undefined & 2\\
\midrule
\multirow{24}[-40]{*}{Psychology, Mental Health and Emotional Support} & Communication Skills \& Empathy & 211\\
 & Child \& Adolescent Mental Health & 199\\
 & Relationship \& Interpersonal Challenges & 181\\
 & Stress, Coping Strategies \& Resilience & 158\\
 & Mood Disorders (Depression \& Bipolar) & 155\\
 & Anxiety, Panic \& Phobias & 112\\
 & Psychological Theories \& Historical Perspectives & 109\\
 & Therapy \& Counseling Methods & 103\\
 & Sexual Orientation, Gender \& Sexual Behaviors & 102\\
 & Trauma \& PTSD & 99\\
 & Emotional Support for Crises \& Suicidal Ideation & 97\\
 & Self-esteem \& Self-sabotage & 95\\
 & Neurodevelopmental Disorders (ADHD, Autism, etc.) & 90\\
 & Addiction \& Substance Use & 69\\
 & Abuse, Violence \& Bullying & 67\\
 & Grief \& Loss & 54\\
 & Personality Disorders & 42\\
 & Schizophrenia \& Psychotic Symptoms & 38\\
 & Social \& Cultural Factors in Mental Health & 37\\
 & Sleep \& Dream Analysis & 36\\
 & Dissociative Disorders \& Maladaptive Daydreaming & 33\\
 & Medication \& Pharmacological Discussions & 28\\
 & Eating \& Body Image Disorders & 25\\
 & Obsessive \& Compulsive Disorders & 16\\
\midrule
\multirow{20}{*}{Interactive Activities with AI Chatbots} & Explicit or Sexual Roleplay & 1023\\
 & Developer Mode or Policy-Breaking Requests & 456\\
 & Interactive Storytelling with User Control & 380\\
 & Comedic or Vulgar Roleplay & 256\\
 & Flirty or Romantic Scenarios & 217\\
 & Childlike or Energetic Roleplay & 188\\
 & Game or Puzzle Interactions & 162\\
 & Roleplay with Personal or Close Relationships & 112\\
 & Fantasy or Mythical Adventures & 101\\
 & Roleplay with Non-Human Traits & 78\\
 & Action or Combat-Based Roleplay & 77\\
 & Testing Chatbot’s Memory or Logic & 68\\
 & Roleplay with Theatrical or Literary Flair & 60\\
 & Roleplay with Real-World Professions & 49\\
 & Minimalistic or Symbolic Responses Only & 44\\
 & Roleplay with Custom Machinery or System Simulation & 43\\
 & Roleplay with Worship or Devotion & 37\\
 & Roleplay with Social or Political Themes & 29\\
 & Roleplay as Rebels or Criminals & 27\\
 & Hypnosis or Therapeutic Roleplay & 7\\
\midrule
\multirow{9}{*}{Linguistics, Language and Translation} & Rewriting and Paraphrasing & 8331\\
 & Translation & 7997\\
 & Vocabulary and Terminology & 2586\\
 & Proof Reading and Grammar Correction & 2102\\
 & Linguistic Analysis & 1099\\
 & Summarization & 779\\
 & Language Learning Assistance & 503\\
 & Phonetics and Pronunciation & 464\\
 & Information Extraction & 391\\
\midrule
\multirow{7}{*}{Social Issues, Politics and Governance} & Domestic Governance \& Public Policy & 1334\\
 & Political Theories \& Ideological Debates & 1231\\
 & International Relations \& Geopolitics & 1190\\
 & Social Justice, Identity \& Cultural Norms & 1009\\
 & Political Leadership \& Electoral Dynamics & 742\\
 & National Security \& Crisis Management & 543\\
 & Economic Policy \& Regulation & 366\\
\midrule
\multirow{27}[-50]{*}{Medicine and Health} & Orthopedics and Musculoskeletal Health & 467\\
 & Nutrition and Dietary Supplements & 466\\
 & Infectious Diseases and Vaccines & 385\\
 & Rehabilitation and Recovery & 384\\
 & Pharmacology and Medication Safety & 378\\
 & Eye, ENT, and Respiratory Conditions & 376\\
 & Surgery and Emergency Care & 341\\
 & Mental Health and Wellbeing & 328\\
 & Reproductive Health and Childbirth & 313\\
 & Digestive, Metabolic, and Endocrine Disorders & 304\\
 & Sexual Health and Function & 243\\
 & Healthcare Systems and Public Health & 238\\
 & Neurology and Nervous System Disorders & 212\\
 & Dermatology and Skin Care & 201\\
 & Diagnostic Tests and Imaging & 190\\
 & Cardiovascular Diseases and Hypertension & 181\\
 & Exercise, Fasting, and Weight Control & 177\\
 & Pediatrics and Child Health & 169\\
 & Preventive Medicine and Wellness & 152\\
 & Cancer and Oncological Care & 141\\
 & Medical Technology and Telemedicine & 109\\
 & Oral Health and Dentistry & 103\\
 & Substance Use and Addiction & 96\\
 & Allergies and Immune Conditions & 88\\
 & Occupational and Environmental Health & 80\\
 & Genetics and Rare Conditions & 76\\
 & Veterinary Medicine and Animal Health & 42\\
\midrule
\multirow{32}{*}{Technology, Engineering and Industry} & Mechanical Engineering and Manufacturing & 678\\
 & Electrical and Electronics Design & 418\\
 & Materials Science and Engineering & 405\\
 & Aerospace and Space Exploration & 381\\
 & Consumer Electronics and Gadgets & 364\\
 & Big Data, IoT, and Smart Systems & 310\\
 & Blockchain and Decentralized Tech & 305\\
 & Networking, Telecommunications, and Cybersecurity & 287\\
 & Civil Engineering and Infrastructure & 278\\
 & Automotive Engineering and Vehicle Technology & 257\\
 & AI and Machine Learning & 251\\
 & VR, AR, and XR Solutions & 245\\
 & Industrial Safety and Compliance & 220\\
 & Robotics, Drones, and Mechatronics & 203\\
 & Military and Defense Technology & 185\\
 & Energy and Sustainable Manufacturing & 156\\
 & Cloud, Virtualization, and Enterprise Platforms & 131\\
 & Supply Chain and Logistics Management & 115\\
 & Software Development and Web Frameworks & 108\\
 & Quantum and High-Performance Computing & 101\\
 & Agricultural Engineering and Food Industry & 84\\
 & Digital Media, Broadcasting, and Streaming & 75\\
 & Hardware Innovation and CPU/GPU Development & 68\\
 & HCI, UI/UX, and Interactive Tech & 67\\
 & Marine and Offshore Engineering & 62\\
 & Data Storage and Retention & 61\\
 & Engineering Education and STEM Training & 55\\
 & Biomedical, Biotech, and Wearables & 55\\
 & Gaming Technology and eSports & 46\\
 & Industrial Digitalization and Change Management & 37\\
 & Product Design and Industrial Innovation & 29\\
 & 3D Printing and Additive Manufacturing & 16\\
\midrule
\multirow{15}{*}{General Digital Support} & AI Capabilities & 472\\
 & AI Limitations & 397\\
 & AI Identity, Version, and Origins & 161\\
 & Correcting or Revising AI Responses & 61\\
 & Technical Guidance: External Apps and Websites & 57\\
 & AI Emotions or Opinions & 48\\
 & Creative Writing & 38\\
 & Official Links or Verification & 33\\
 & Coding Tasks & 29\\
 & Technical Guidance: Phones and Software & 24\\
 & Email and Account Management & 19\\
 & Comparison with Other AI Systems & 18\\
 & Education or Research Use & 17\\
 & Search and Browsing Advice & 10\\
 & Payment or Subscription & 5\\
\midrule
\multirow{5}{*}{Food, Cooking and Nutrition} & Nutritional Guidance \& Diet Planning & 569\\
 & Recipes \& Cooking Techniques & 518\\
 & Ingredient Selection \& Quality & 218\\
 & Culinary Culture \& Dining Experience & 166\\
 & Food Safety \& Storage & 76\\
\midrule
\multirow{13}{*}{Art and Design} & Product \& Merchandise Design & 1086\\
 & AI-Generated Art \& Prompt Engineering & 585\\
 & Digital Media \& Advertising Design & 492\\
 & Color Theory \& Visual Composition & 407\\
 & Character \& Animation Design & 290\\
 & Art History \& Critique & 270\\
 & Editorial \& Commercial Illustration & 262\\
 & Fashion \& Costume Design & 252\\
 & Logo \& Branding Design & 213\\
 & Educational \& Children's Art & 204\\
 & Architectural \& Environmental Design & 192\\
 & Digital Art \& Software Techniques & 132\\
 & Traditional \& Manual Art Techniques & 116\\
\midrule
\multirow{10}{*}{Religion, Mythology and Spirituality} & Biblical and Scriptural Narratives & 981\\
 & Islamic Sacred Narratives & 363\\
 & Classical Mythology Narratives & 356\\
 & Eastern Sacred Narratives & 243\\
 & Modern Esoteric and Occult Spirituality & 188\\
 & Religion, Society, and Cultural Critique & 178\\
 & Astrological and Divinatory Traditions & 169\\
 & Folk and Indigenous Myth Narratives & 164\\
 & Norse and Germanic Mythological Narratives & 44\\
 & Ancient Near Eastern and Persian Narratives & 31\\
\midrule
\multirow{4}{*}{Literature and Book Analysis} & Narrative and Prose Analysis & 1482\\
 & Poetry and Versified Analysis & 427\\
 & Literary Guidance and Recommendations & 355\\
 & Advanced Literary Criticism & 43\\
\midrule
\multirow{18}{*}{Philosophy and Ethics} & Epistemology, Logic, and Fallacies & 349\\
 & Law, Governance, and Political Philosophy & 341\\
 & Mind, Consciousness, and Reality & 303\\
 & Religion, Theology, and Faith Traditions & 299\\
 & Existentialism, Death, and Meaning & 176\\
 & Moral Theories, Virtue, and Character Development & 171\\
 & Moral Speech and Expression & 146\\
 & Critical Theory and Postmodernism & 133\\
 & Consent, Power, and Manipulation & 104\\
 & Cultural Norms and Social Ethics & 100\\
 & Aesthetics and Artistic Philosophy & 91\\
 & Ethics in AI and Future Technologies & 90\\
 & Professional Ethics and Duty & 81\\
 & Markets, Capitalism, and Economic Fairness & 43\\
 & Bioethics, Medicine, and Life Origins & 42\\
 & Morality Toward Animals & 40\\
 & Love, Relationships, and Emotional Ethics & 28\\
 & Environmental Ethics and Sustainability & 19\\
\midrule
\multirow{19}[-20]{*}{Sports and Athletics} & NCAA College Football & 1012\\
 & Motorsport & 607\\
 & NBA Basketball & 604\\
 & NCAA College Basketball & 549\\
 & Global Soccer & 538\\
 & Fictional or Hypothetical Scenarios & 451\\
 & Professional American Football & 313\\
 & General or Cross-Sport Training \& Fitness & 218\\
 & Professional Wrestling & 146\\
 & Baseball & 68\\
 & Combat Sports & 64\\
 & Cricket & 60\\
 & Cycling (Races \& Gear) & 59\\
 & Ice Hockey & 25\\
 & Tennis and Other Racket Sports & 18\\
 & Rugby & 14\\
 & Gymnastics \& Swimming & 7\\
 & Volleyball & 3\\
 & Golf & 2\\
\midrule
\multirow{33}{*}{Environment, Ecology and Sustainability} & Climate Change Causes, Impacts, and Adaptation & 140\\
 & Biodiversity Conservation and Wildlife Protection & 119\\
 & Greenhouse Gas Emissions and Carbon Management & 117\\
 & Pollution (Air, Water, Soil) and Remediation & 102\\
 & Waste Management and Circular Economy & 101\\
 & Environmental Policies, Laws, and Regulations & 82\\
 & Sustainable Energy and Energy Transition & 74\\
 & Green Industry, Corporate Sustainability, and Innovation & 72\\
 & Water Resource Management and Conservation & 67\\
 & Ecological Economics and Sustainable Development & 66\\
 & Environmental Education and Public Awareness & 45\\
 & Deforestation, Reforestation, and Sustainable Forestry & 43\\
 & Environmental Monitoring, Data Analysis, and Reporting & 40\\
 & Sustainable Lifestyles and Consumer Choices & 39\\
 & Sustainable Packaging, Recycling, and Plastics Reduction & 37\\
 & Sustainable Agriculture and Food Systems & 35\\
 & Marine and Coastal Conservation & 33\\
 & Sustainable Cities and Urban Development & 33\\
 & Ecological Restoration and Ecosystem Management & 33\\
 & Digital Technologies and Sustainability & 32\\
 & Sustainable Architecture and Construction & 26\\
 & Sustainable Transportation and Mobility & 23\\
 & Soil Health and Land Use Management & 22\\
 & Environmental Disaster Preparedness and Risk Reduction & 20\\
 & Carbon Markets and Climate Finance & 19\\
 & Eco-friendly Materials and Green Design & 17\\
 & Community-based Conservation and Participation & 15\\
 & Climate Negotiations and International Agreements & 12\\
 & Protected Areas and Natural Heritage Sites & 12\\
 & Environmental and Climate Justice & 11\\
 & Conservation Technology and Innovation & 6\\
 & Environmental Impact Assessment and Life Cycle Analysis & 5\\
 & Sustainable Tourism and Ecotourism & 3\\
\midrule
\multirow{10}{*}{Travel and Tourism} & Cultural, Heritage \& City Experiences & 126\\
 & Transport \& Logistics & 87\\
 & Travel Itineraries \& Trip Planning & 65\\
 & Accommodation \& Lodging & 54\\
 & Tourism Industry, Policy \& Market & 49\\
 & Culinary \& Dining & 40\\
 & Visa \& Travel Documentation & 40\\
 & Beach, Coastal \& Cruise Tourism & 37\\
 & Entertainment \& Nightlife & 28\\
 & Adventure \& Outdoor Activities & 25\\
\midrule
\multirow{15}[-30]{*}{Professional Development and Career Advice} & Cover Letters \& SOPs & 270\\
 & Resume \& CV Enhancement & 233\\
 & Workplace Culture \& Dynamics & 132\\
 & Skill Development \& Advanced Education & 128\\
 & Leadership \& Team Management & 106\\
 & Salary \& Compensation Guidance & 96\\
 & Recruitment \& Talent Acquisition & 96\\
 & Industry-Specific Career Advice & 75\\
 & LinkedIn \& Personal Branding & 69\\
 & Job Search \& Networking Strategies & 60\\
 & Career Transitions \& Upskilling & 60\\
 & Negotiation \& Employment Contracts & 42\\
 & Interview Preparation \& Techniques & 31\\
 & Employment Documentation \& Verification & 31\\
 & Freelancing \& Entrepreneurship & 19\\
\midrule
\multirow{17}{*}{Home and Household} & Gardening: Planting \& General Care & 140\\
 & Gardening: Soil \& Fertilization & 128\\
 & Fruit \& Berry Cultivation & 107\\
 & Home Fixtures \& Materials & 83\\
 & Gardening: Pest \& Disease Management & 75\\
 & Interior Design \& Decoration & 60\\
 & Home Maintenance \& Appliance Repair & 54\\
 & Laundry \& Fabric Care & 36\\
 & DIY Tools \& Household Projects & 31\\
 & Household Cleaning \& Stain Removal & 27\\
 & Outdoor Landscaping \& Mulching & 24\\
 & Eco-Friendly \& Sustainable Practices & 15\\
 & Household Safety \& Security & 14\\
 & Real Estate \& Tenancy & 13\\
 & Household Management \& Lifestyle & 13\\
 & Home Organization \& Storage Solutions & 8\\
 & Household Pets \& Animal Care & 5\\
\end{longtable}
\twocolumn
\clearpage

\normalsize            

\begin{figure*}[!t]
    \centering
    \includegraphics[width=\linewidth]{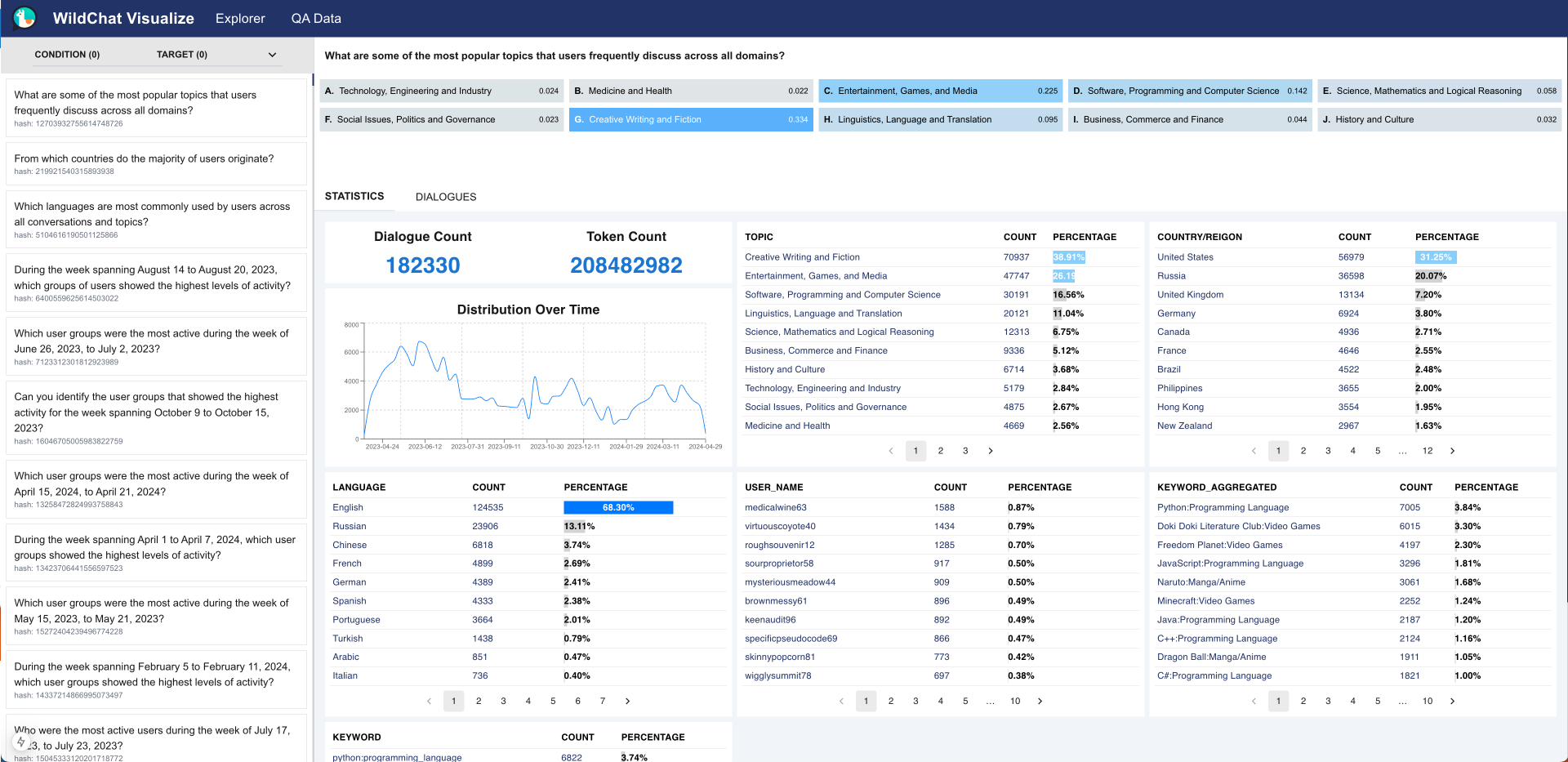}
    \caption{Data Visualization Demo Overview}
    \label{fig:demo-main}
\end{figure*}

\vspace{1em} 

\section{Data Visualization Demonstration}

\begin{figure}[h]
    \centering
    \includegraphics[width=0.95\linewidth]{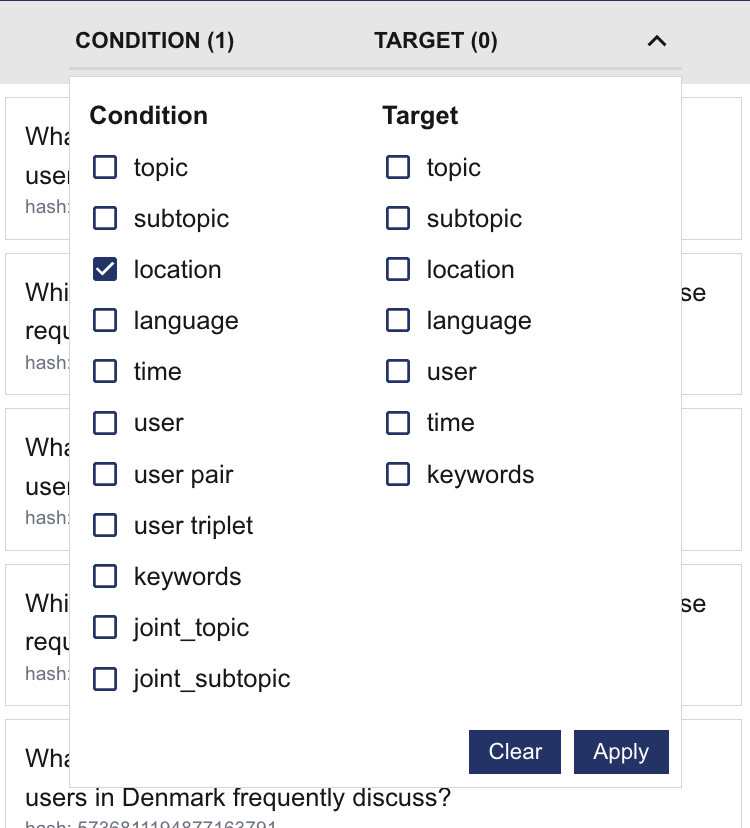}
    \caption{Question filter attributes of different conditions and targets.}
    \label{fig:demo-filter}
\end{figure}

We developed an interactive data visualization interface using React.js and Next.js for the frontend, and FastAPI for the backend implementation. MongoDB serves as the database system. An overview of the interface is shown in \Cref{fig:demo-main}. Users can filter generated questions using a configurable question filter, as illustrated in \Cref{fig:demo-filter}.

The filtering mechanism allows users to select one or more attributes for both the condition and target fields to retrieve relevant questions. For instance, the filters ``user\_pair'' and ``user\_triplet'' refer to questions based on common interests between two or three users, respectively. Similarly, ``joint\_topic'' and ``joint\_subtopic'' denote filters that select conversations involving shared topics or subtopics.

\begin{figure}[h]
    \centering
    \includegraphics[width=0.95\linewidth]{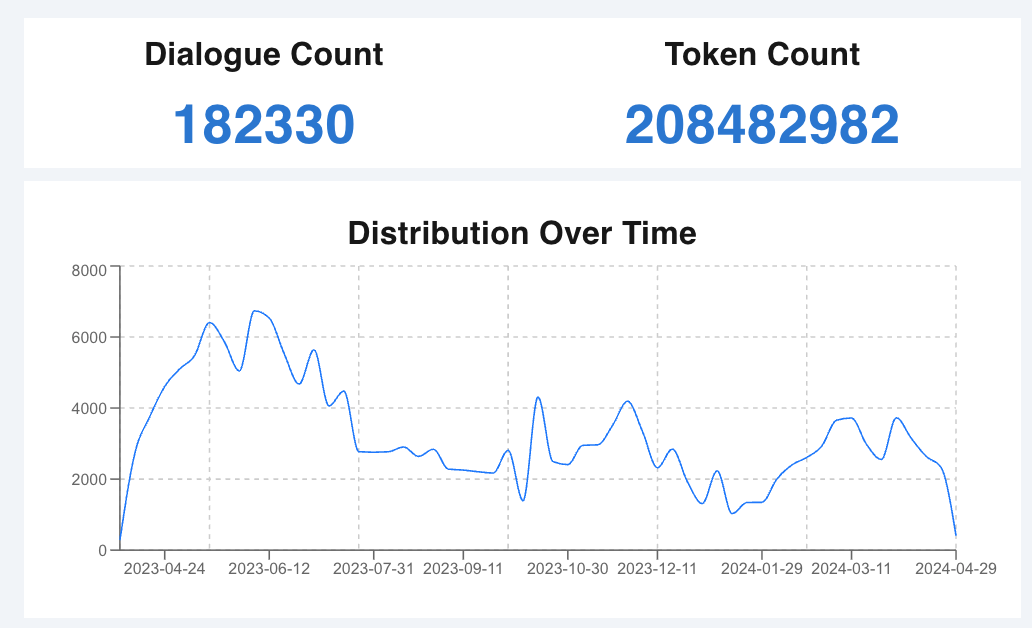}
    \caption{Context conversation and token count and distribution of conversation over time. }
    \label{fig:demo-filter2}
\end{figure}

\begin{figure}[h]
    \centering
    \includegraphics[width=0.95\linewidth]{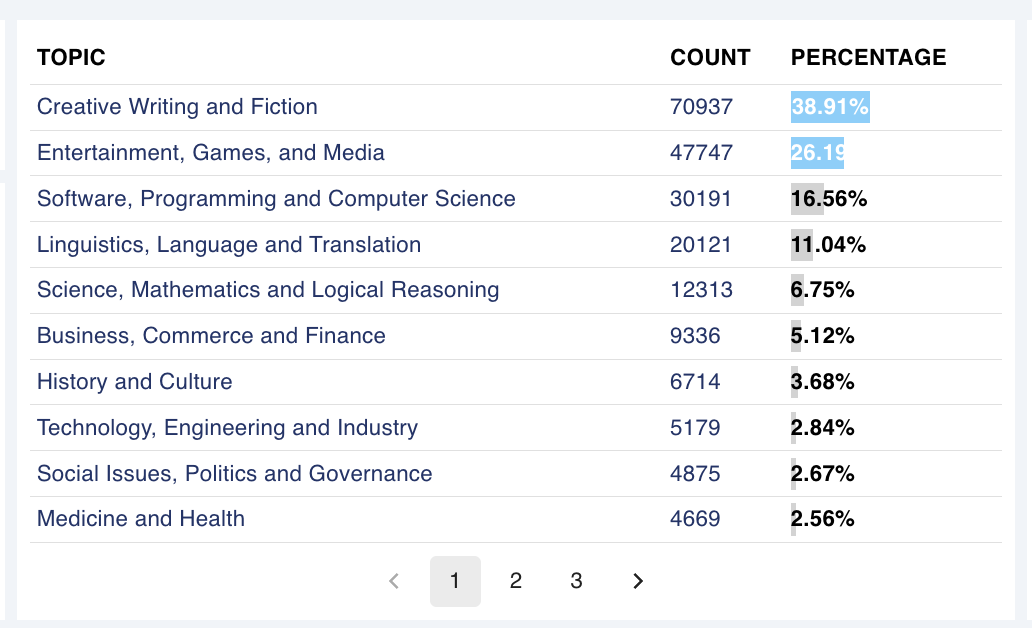}
    \caption{Distribution of topics}
    \label{fig:demo-filter3}
\end{figure}

For each question, the interface displays the number of supporting dialogues and their associated token counts. Additional distributions—such as raw keywords, aggregated keywords, language, topic, location, and user identity—are visualized to facilitate deeper insights.

\begin{figure*}[ht]
    \centering
    \includegraphics[width=0.95\linewidth]{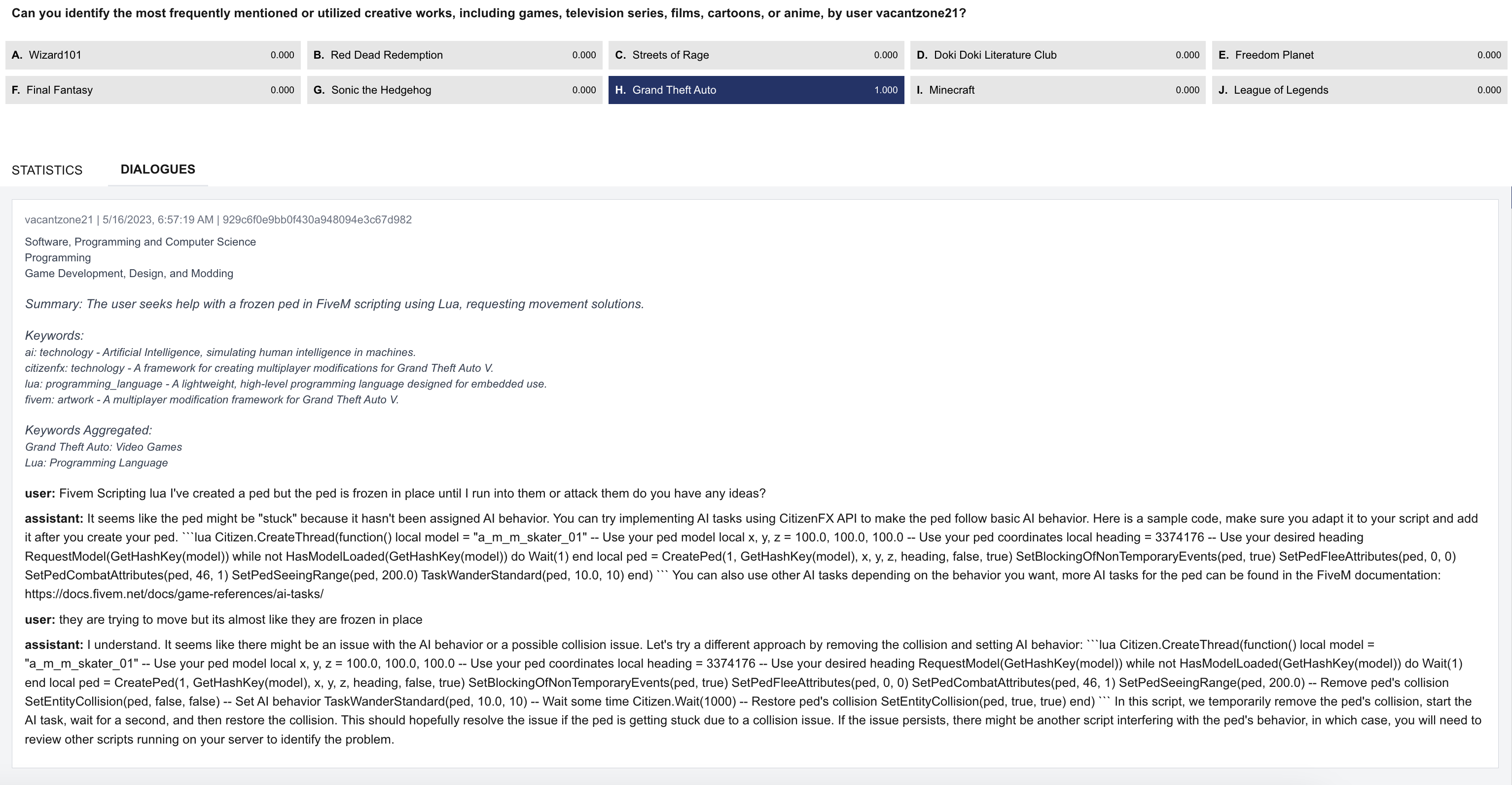}
    \caption{Dialogue Detail Display}
    \label{fig:demo-main2}
\end{figure*}

Users can also explore the ``DIALOGUES'' panel to view all conversation excerpts that support a particular question. Each dialogue entry includes detailed metadata: username, timestamp, topic, subtopic, generated summary, raw extracted keywords, and aggregated keywords. This comprehensive display allows users to audit or explore the basis of each proposed question in context.

\section{Experiment Implementation Details}

\label{appendix:implementation_details}

We employed MongoDB v8.0.4 for question proposal generation and ground-truth-based retrieval. All retrieval experiments utilizing BM25 and dense kNN methods were conducted using Elasticsearch v8.18. Training and inference for open-source models were carried out on a range of GPUs, including the NVIDIA RTX A6000 Ada, NVIDIA H100, and NVIDIA H200, depending on availability.

For all embedding-based dense retrieval experiments, the questions, generated queries, documents, and summaries were encoded using the OpenAI \texttt{text-embedding-3-large} model, which produces 3072-dimensional vectors.

For fine-tuning experiments with Qwen3-8B, we used the HuggingFace Transformers library \cite{wolf2020huggingfacestransformersstateoftheartnatural}, version 4.51.3, training on the full conversation dataset with a peak learning rate of $1 \times 10^{-5}$, a batch size of 8, and a linear learning rate decay schedule.

For text pre-processing of RAG, we chunked raw conversations into segments of at most 512 tokens with a maximum overlap of 128 tokens, preserving sentence boundaries wherever possible. For summarized conversations, no chunking is performed due to relatively short text length.

For inference with open-source models, we utilized vLLM v0.8.5.post1. The sampling hyperparameters used during inference are detailed in \Cref{tab:inference_param}.

\begin{table}[h]
    \centering
    \small
    \begin{tabular}{lccc}
        \toprule
        Model Name      & top\_p    & top\_k    &   temperature \\
        \midrule
        Gemma3-4B       & 0.95      &  64       &   1.0         \\
        Qwen3-8B        & 0.8       &  20       &   0.7         \\
        Qwen3-8B-Think  & 0.95      &  20       &   0.6         \\
        Qwen3-32B       & 0.8       &  20       &   0.7         \\
        Qwen3-32B Think & 0.95      &  20       &   0.6         \\
        GPT-4.1-mini    & 1.0       & -         &   1.0         \\
        o4-mini         & -         & -         &   -           \\
        \bottomrule
    \end{tabular}
    \caption{Model sampling hyper-parameter}
    \label{tab:inference_param}
\end{table}

For broad query generation in \textbf{PROBE}, we use GPT-4.1-mini as query and filter generator with top\_p = 0.5 and top\_k = 0.5.

\definecolor{borderblue}{RGB}{34,51,103}    
\definecolor{bggray}{RGB}{245,247,250}

\section{Data Construction Process}
\label{appendix:data_creation}
In this part, we explain in detail how we create the dataset. We start with WildChat-Full dataset which contains around 990K conversations. 

\subsection{Pre-processing and De-duplication}
We begin by de-duplicating the full WildChat dataset using MinHash and Locality-Sensitive Hashing (LSH), following the approach described in \citet{huggingface_dedup_2023}. For MinHash, we use 4-grams ($k = 4$) and 9 permutations ($p = 9$). For LSH, we set the band size to $b = 7$ and the row size to $r = 3$. After de-duplication, approximately 520K conversations remain.

Next, we tokenize all conversations using the LLaMA 3 tokenizer \cite{grattafiori2024llama3herdmodels} and discard those exceeding 4,096 tokens. Users are identified based on a combination of hashed IP addresses and HTTP request headers, and each user is assigned a randomized username. Users with fewer than 10 sessions are considered inactive, and all their conversations are removed.

After filtering by conversation length and user activity, around 220K conversations remain. All subsequent processing steps are performed on this filtered dataset.

\begin{figure*}[htbp]
    \centering
   \begin{tcolorbox}[colframe=borderblue, colback=bggray, left=1mm, right=1mm, top=0.5mm, bottom=0.5mm]
\begin{minted}{markdown}
# Context

You are a helpful assistant in processing data. You are going to generate a report for a user chatbot interaction dialogue.

In the data given below, user requests starts with [**User Request**] and agent response starts wtih [**Agent Reponse**]. Utterance are separated by '----'. 

# Content

{{input_text}}

# Instruction 

You need to generate report satisfying following requirements based on Content:

1. Extract or infer all keywords of following types from the dialogue:

    - person: individuals’ names, including first, middle, and last names, titles, and honorifics. Example: Nelson Mandela, Dr. Jane Doe
    - technology: Terms describing technology of any fields. Example: AI, 5G, renewable enery, NFT, SEO, Large Language Model, AR, VR, Metaverse.
    - scientific_term: Terms describing science theories, or concepts. Example: Quantum Physics, Photosynthesis
    - food: Food-related terms, ingredients, or dishes. Example: Avocado, Chocolate.
    - demographic_term: term references to ethnicities, nationalities, or demographic groups. Example: LGBTQ+, Caucasian, Afican American.
    - organization: Companies, institutions, government agencies, and other organized groups. Example: Google, Meta, United Nations, World Health Organization, MIT, Stanford, FDA.
    - location: Geographical locations, including stars, planets, countries, cities, states, addresses, and landmarks. Example: London, Mount Everest, Times Square, United States, Moon, Neptune, Sun.
    - event: Name of social, cultural, military, political, historical, scientific, commercial, religious, medical or health events. Example: World War II, 2024 Paris Olympic, Cold War, CES 2024, Industrial Revolution, The Renaissance. 
    - artwork: Name of any form artworks, including music, books, video games, anime, comic, drama, shows, TV shows, TV series, films, painting etc.
    - programming_language: Any kind of programming language. Example: Python, Java, C++, C#, LaTeX, R, CSS etc.
    - product_brands: Name of products and brands. Example: IPhone 14, Nike Air Max, Apple Mac Book.
    - financial_term: financial or economic terminology. Example: Interest Rate, Inflation. 


2. DO NOT output "none" if specfic kind of keywords not appear.

3. The keywords extracted MUST be **uniquely idenfiable without context**.

4. Give simple description of each keywords **within 15 words** in **English**.

5. All keywords extracted MUST be **English** or translated into **English**.

6. Write a summary of given user chatbot interaction **within 30 words** in **English**, focus on user query, describe from third person view.

7. Keep as much information as possible in summary about user request.

8. Explain user's intent based on the given content, respond in `intent` part within **30 word** using **English**.

9. The answer MUST be generated in json format:
    {
        "summary": "<summary>",
        "intent": "<intent>",
        "keywords":[
            {
                "keyword_type": <type_1>,
                "value": <value_1>,
                "description": <keyword_description_1>
            },
            {
                "keyword_type": <type_2>,
                "value": <value_2>,
                "description": <keyword_description_2>
            },
        ],
    }

# Response
\end{minted}
\end{tcolorbox}
    \caption{Prompt for keywords extraction and summarization. }
    \label{fig:prompt_extraction}
\end{figure*}

\begin{algorithm*}[htbp]
\caption{TnT-LLM: Taxonomy Generation Phase}
\label{alg:tnt-llm-taxonomy-topic}
\begin{algorithmic}[1]
\Statex \textbf{Input:} Max round of iteration $N$, Batch size $B$, Conversations summaries $C$, Summary embeddings $E$, 2Number of cluster of KMeans $K$, Initial taxonomy generation prompt $P_{\text{initial, topic}}$, Taxonomy update prompt $P_{\text{update, topic}}$
\Statex \textbf{Output:} Label taxonomy $T$

\State Partition summaries $C$ into $K$ clusters $\{D_1, \dots, D_K\}$ using KMeans on $E$.
\State Initialize taxonomy $T \leftarrow \emptyset$.
\State Initialize cursors for round-robin sampling from each cluster $D_k$.

\For{$n \leftarrow 1 \text{ to } N$}
    \State $S_{batch} \leftarrow \emptyset$
    \State Select up to $B$ summaries for $S_{batch}$ by sampling from clusters $\{D_k\}$ in a round-robin fashion without replacement, advancing cursors.
    
    \If{$S_{batch}$ is empty} \Comment{No more summaries available for sampling}
        \State \textbf{break}
    \EndIf

    \If{$n = 1$}
        \State $T \leftarrow \text{CallLLM}(P_{\text{initial, topic}}, S_{batch})$
    \Else
        \State $T, \text{score} \leftarrow \text{CallLLM}(P_{\text{update, topic}}, S_{batch}, T)$ \Comment{Update existing $T$}
    \EndIf

    \If{score not improve for 3 iteration}
        \State \textbf{break}
    \EndIf
\EndFor
\State \Return $T$

\end{algorithmic}
\end{algorithm*}

\subsection{LLM-based keywords and summarization extraction}

To perform TnT-LLM for topic discovery, we begin by extracting keywords and summaries from raw conversations. Specifically, we prompt GPT-4o to generate both the keyword set and a concise summarization of each conversation. The extracted keywords span a diverse set of semantic types, including persons, technologies, scientific terms, foods, demographic terms, organizations, locations, events, artworks, programming languages, product brands, and financial terms. The complete prompt used for this extraction process is shown in \Cref{fig:prompt_extraction}.

\begin{figure*}[htbp]
    \centering
   \begin{tcolorbox}[colframe=borderblue, colback=bggray, left=1mm, right=1mm, top=0.5mm, bottom=0.5mm]
\begin{minted}{markdown}

# Context

You are a helpful assistant for clustering human-AI conversation. The following content are a batch of human-AI conversation summary sampled, separated by "----". You are going to propose a set of meaningful, diverse and high quality categories so that all human-AI conversation can be classified without ambiguity.

# Content

{{input_text}}


# Instruction 

Your task is to propose a list classes and corresponding description so that the given data can be classified into, with following requirements:

1. The classes generated are the **domain** of human-AI interaction, avoid introducing user intent.

2. The class names and class descriptions generated can **accurately** and **consistently** classify new data points **without ambiguity**.

4. The class name should be a **concise and clear label** for the category.

5. The classes generated MUST be **mutual exclusive**.

6. The class description of each class should be generated within **100 words** in English.

7. The class name and class description must be consistent with each other.

8. Output class must match the data as close as possible, without adding unnecessary ones and missing necessary ones.

9.  Generate **No More Than 30 classes**

10. Avoid categories include any vague information such as "Other", "Undefined", "Miscellaneous". 

11. The response should be generated in json format following:

    {
        "classes": [
            {
                "class_description" : <description_1>,
                "class_name" : <title_1>
            },
            {
                "class_description" : <description_2>,
                "class_name" : <title_2>
            },
            {
                "class_description" : <description_3>,
                "class_name" : <title_3>
            },
             <more classes...>
        ]
    }

Make sure output **pure json**

# Response
\end{minted}
\end{tcolorbox}
    \caption{Initial Taxonomy Generation Prompt}
    \label{fig:topic_initial}
\end{figure*}

\begin{figure*}[htbp]
    \centering
   \begin{tcolorbox}[colframe=borderblue, colback=bggray, left=1mm, right=1mm, top=0.5mm, bottom=0.5mm]
\begin{minted}{markdown}
# Context

You are a helpful assistant for clustering human-AI conversation. The following content in **Content** part are a batch of human-AI conversation summary sampled, separated by "----". And a category table you generaeted based on the previous data in **Category Table** part. You are going to update the table for downstream user interest discovery.

# Category Table

{{input_category_table}}

# Content

{{input_text}}

# Requirements

Your need to update the category table to make sure the table satisfy the following **requirements**:
    - The classes generated are the **domain** of human-AI interaction, avoid introducing user intent.
    
\end{minted}
\end{tcolorbox}
\end{figure*}

\begin{figure*}[htbp]
    \centering
   \begin{tcolorbox}[colframe=borderblue, colback=bggray, left=1mm, right=1mm, top=0.5mm, bottom=0.5mm]
\begin{minted}{markdown}

    - The class names and class descriptions generated can **accurately** and **consistently** classify new data points **without ambiguity**.
    
    - The class name should be a **concise and clear label** for the category.
    
    - The classes generated MUST be **mutual exclusive**.
    
    - The class description of each class should be generated within **100 words** in English
    
    - The class name and class description must be consistent with each other.
    
    - Output class must match the data as close as possible, without adding unnecessary ones and missing necessary ones.
    
    - The generated classes must useful for user interest discovery and analysis.

    - Generate **No More Than 60 classes**

    - Avoid including three or more different aspects in one category, such as `History, Politics & Government`. 

    - Avoid categories include any vague information such as "Other", "Undefined", "Miscellaneous". 

# Instructions

You need to update using following steps: 

1. Review the given category table and the input data. Provide a rating score of current table. The rating score should between 0 to 100. The score should be given based instrinstic quality and extrinstic quality:

    - **Instrinstic quality**
        1) If the categories meets the requirements given in **Requirements** part, with clear and consistant category names and descriptions, and no overlap or contracdiction among the categories.
        2) If the categories include any vague information such as "Other", "Undefined", "Miscellaneous". 
        3) If there is categories that are too general and include too many aspects or sub-categories.

    - **Extrinstic quality**
        1) If the data given can be classified into the given category consistently without any ambiguity. 
        2) If there is missing category that the data can not classified into.
        3) If there is any category that is unnecessary so that can be merged or removed.

2. Based on your score, decide if you need to update the categories, you can perform following operations:
    - Edit class name or class description of the categories.
    - Add new categories if there are missing categories.
    - Split one categories into multiple to become specific.
    - Merge multiple categories into one to become less amiguous. 
    - Remove unnecessary categories to reduce redundency.
    - No update if they are good enough.
    
    If you decide to update the categories, explain the update sugguestion in `suggesion` part. Otherwise just output `N/A` in suggestion part.

    Restate: The categories should be **concise, consistent, mutual exclusive**. Make sure remember to update the dialogue count accordingly.

    Restate: Be **specific** about each category. **Do not include vague categories**

    You can ignore low quality or ambuiguous data points.


4. Output the report using json format as follows based on your decision and review result above, make sure categories satisfy the **requirements** given. 
    {
        "score": <table_score>,
        "suggestion: <suggestion>,
        "classes": [
            {
                "class_description" : <description_1>,
                "class_name" : <title_1>
            },
            {
                "class_description" : <description_2>,
                "class_name" : <title_2>
            },
            {
                "class_description" : <description_3>,
                "class_name" : <title_3>
            },
             <more classes...>
        ]
    }

# Updated Category Table
\end{minted}
\end{tcolorbox}
    \caption{Taxonomy Update Prompt}
    \label{fig:topic_update}
\end{figure*}

\subsection{TnT-LLM based Topic and Subtopic Discovery and Assignment}
\subsubsection{Topic Discovery and Assignment}
\paragraph{Topic Taxonomy Generation}
We largely follow the pipeline of TnT-LLM \cite{wan2024tntllmtextminingscale} to identify topics within the dataset. Rather than randomly sampling from a large corpus, we first obtain the textual embeddings of conversation summaries using the \texttt{BAAI/bge-en-icl} model \cite{li2024makingtextembeddersfewshot}. We then perform clustering on these embeddings to guide our sampling, ensuring a diverse selection across different semantic regions. This step is added to enhance topic diversity in the sampled subset.

Subsequently, we apply the topic discovery algorithm detailed in \Cref{alg:tnt-llm-taxonomy-topic}. The initial taxonomy generated is visualized in \Cref{fig:topic_initial}, while the prompt used for topic refinement is shown in \Cref{fig:topic_update}. For all topic discovery steps, we employ GPT-4o as the underlying language model, using hyperparameters $B=K=500$ and $N=10$. To perform efficient KMeans clustering, we utilize the FAISS library \cite{douze2025faisslibrary}. Unlike the original TnT-LLM method, which relies on LLMs for taxonomy refinement, we manually resolve conflicts and enforce mutual exclusivity among the discovered topics.

\begin{figure*}[htbp]
    \centering
   \begin{tcolorbox}[colframe=borderblue, colback=bggray, left=1mm, right=1mm, top=0.5mm, bottom=0.5mm]
\begin{minted}{markdown}
# Context

You are a helpful assistant in analyzing user-AI interaction data. You are going to classify a user-AI interaction conversation based on a category table. The **Content** and **Categories** are given in json format.

In the data given below, user requests starts with <User Request> and agent response starts wtih <Agent Response>. Utterance are separated by '----'. 

# Content

{{input_text}}

# Categories

{{input_categories}}


# Classification Examples

You need to labeling based on user request or demand, here are some examples, separated by `----`:

{{examples}}

# Instruction 

You need to classify the given conversation using the `conversation`, `summary`, # Categories table and given # Classification Examples with following requirements:

- Explain how you perform the classification in `explanation` part **WITHIN 200 WORDS**.

- `Entertainment, Games, and Media` MUST be added with proper relevance order if there are **LESS THAN THREE** other classes **AND** the **MAJOR** characters, content, plot, universe, celebrities involved in conversation is from a known game, film, tv series, comics or other artwork for entertainment described in #Categories.

- `Erotic, Explicit and Inappropriate Content` MUST be ranked LOWEST if **EXPLICITLY INVOLVED**. 

- Classify based on the <User Request> in `conversation`, then refer to <Agent Response>, finally refer to `summary` if necessary.

- You must classifiy the conversation into **AT MOST THREE** classes **MOSTLY RELEVANT**.

- The classification result MUST have **AS SMALL NUMBER OF CLASS AS POSSIBLE**. 

- AVOID classify the conversation into categories that slightly involved, and focus on users' **MAJOR DEMAND**.

- Respond the classes **ORDER BY RELEVANCE**.

- All response should be in **ENGLISH**

- The classification MUST be done based on `class_description`, `class_examples` and # Classification Examples.

- Respond in **pure json** following with explanation and selected class index:
    {
        "explanation": <explanation>,
        "classes": [<class_index_1>, <class_index_2> ...]
    }

# Response

\end{minted}
\end{tcolorbox}
    \caption{Topic Assignment Prompt}
    \label{fig:topic_assignment}
\end{figure*}

\paragraph{Topic Label Assignment}  Using the generated topics and corresponding taxonomy, we assign a topic ID to each conversation. This assignment process can be formulated as a multi-label classification task. The labeling is performed by GPT-4o using the assignment prompt illustrated in \Cref{fig:topic_assignment}. The prompt is carefully designed to mitigate common errors identified through a manual inspection of a small validation set consisting of 400 examples.


\subsubsection{Subtopic Discovery and  Assignment}
\begin{figure*}[htbp]
    \centering
   \begin{tcolorbox}[colframe=borderblue, colback=bggray, left=1mm, right=1mm, top=0.5mm, bottom=0.5mm]
\begin{minted}{markdown}
You are an expert in analyzing and summarizing dialgoue between user and chatbot, you are going to summarize following conversation based on instruction. 

{{conversation}}

# Instructions

- You need to summarize the dialogue between user and ai chatbot from {{class_name}} topic aspect, the **definition** of the topic is:
    {{class_description}}

- You MUST check if the conversation contains user request or input related to {{class_name}} based on the **definition**, explain your check result briefly within 50 words. 

- The check result MUST be either "yes" or "no", a string in lower case.

- You need to keep as much information as possible, try your best to keep important keywords and facts in the dialogue.

- The summary MUST describe from third person perspective and **focus on user request**.

- The summary MUST be done within 10 - 20 words using one sentence related to {{class_name}}.

- Make the summary a perfect version for sub-topic discovery.

- Respond in following format using **pure json**
    
    {
        "explanation": "<explanation>",
        "check_result": "<check_result>",
        "summary": "<summary>"
    }

# Response 
\end{minted}
\end{tcolorbox}
    \caption{Topic Validation and Aspected Summarize Prompt}
    \label{fig:subtopic_aspected_summarize}
\end{figure*}
\paragraph{Subtopic Taxonomy Generation} For each discovered topic, we further identify its subtopics by running TnT-LLM on all conversations classified under that topic. However, subtopic discovery proves to be more challenging. To address this, we adopt a more sophisticated pipeline and employ a stronger model. The following pipeline is specifically designed to facilitate subtopic discovery within each major topic.

\begin{itemize}
    \item[1.] Prompt GPT-4o to check the result of topic assignment and summarize the raw conversation from the perspective of major topic using the prompt shown in \Cref{fig:subtopic_aspected_summarize}. 
    \item[2.] Get the embedding of the summaries that pass checking using \texttt{text-embedding-3-large}.
    \item[3.] Run KMeans use faiss with $K$ in $\{10, 15, 20, 25, 30, 35, 40\}$, find the top 3 best number of centroid $k_1^*, k_2^*, k_3^*$ using silhouette score \cite{ROUSSEEUW198753}. 
    \item[4.] For each target number of subtopics $k^*$,we execute \Cref{alg:tnt-llm-taxonomy-topic} with parameters $B = 200, K=200, N = 30$ using topic-specific initial and update prompts as illustrated in \Cref{fig:subtopic_initial} and \Cref{fig:subtopic_update}. The model used for subtopic discovery is OpenAI-o1, selected for its strong reasoning capabilities. 
    To enforce the desired number of generated subtopics at the start of the iteration, we replace the placeholder ``\{min\_class\_number\_requirement\}'' in \Cref{fig:subtopic_initial} with instruction ``- Generate NO LESS THAN $k^*$ topics.'' . 

    \item [5.]After generating the taxonomy for each $k^*$, we randomly sample 10\% of data instances from the current topic—capped at a maximum of 1000 samples. We then query the o3-mini model, which has strong reasoning ability, using the prompt provided in \Cref{fig:subtopic_assignment}. This yields a set of predicted labels $\{l_1, l_2, \cdots, l_i, \cdots, l_m\}$, along with corresponding relevance scores $\{r_1, r_2, \cdots, r_i, \cdots, r_m\}$ between 0-10, each ranging from 0 to 10. We then compute a quality score for each generated taxonomy using the following equations:

    \begin{equation}
        s_{\text{quality}} = s_{\text{coverage}} + s_{\text{certainty}}
    \end{equation}

    Where $s_{\text{coverage}}$ and $s_{\text{certainty}}$ are defined as: 
    \begin{equation}
        s_{\text{coverage}} = 1.0 - \frac{N_{\text{Undefined}}}{N}
    \end{equation}
    
    where $N_{\text{Undefined}}$ is the number of samples that labeled as ``Undefined'', which is not fit in the taxonomy, and $N$ is the number of data sample labeled for taxonomy validation.
    \begin{equation}
    \begin{aligned}
        p_i &= \frac{r_i}{\sum_{k=0}^m r_k} \\
        H_j &= \frac{\sum_{i=1}^{n}p_i\log_2 p_i}{\log_2m} \\
        s_{\text{certainty}} &=   \frac{\sum_{j=1}^N (1.0 - H_j)} {N}
    \end{aligned}
    \end{equation} 

    We select the best taxonomy generated using $s_{\text{quality}}$. 
 \end{itemize}

\begin{figure*}[htbp]
    \centering
   \begin{tcolorbox}[colframe=borderblue, colback=bggray, left=1mm, right=1mm, top=0.5mm, bottom=0.5mm]
\begin{minted}{markdown}
# Context

You are a helpful assistant for clustering human-AI conversation within topic "{{topic}}". The following # Input Data are a batch of summarized human-AI conversation sampled. You are going to propose a set of meaningful, diverse and high quality categories so that all human-AI conversation can be classified without ambiguity.

# Input Data

{{input_text}}

# Instruction 

Your task is to propose a list sub-topic within topic of {{topic}} and corresponding description so that the given data can be classified into, with following requirements:
    - The classes generated are the **TOPIC** MUST fall under the parent topic "{{topic}}". 
    - The parent **topic description** are as follows:
        {{topic_description}}
    - The class names and class descriptions generated can **ACCUREATELY** and **CONSISTENTLY** classify new data points into **1-3 class** with **NO AMBIGUITY**.
    - The class name should be a **CONCISE AND CLEAR** short sentence for the category.
    - The classes generated MUST be **MUTUAL EXCLUSIVE**.
    - The class description of each class should be generated within **200 WORDS** in English.
    - The class description MUST be generated based on data sample.
    - The class name must be consistent with its class description.
    - Output class must **fit the data as close as possible**, avoid adding unnecessary ones and missing necessary ones.
    - Avoid general categories include any vague information such as "Other Topics", "Undefined", "Miscellaneous". 
    - You may ignore data points not related to {{topic}}. 
    - Keep each class **fine-grained**, AVOID include too many aspect in one class.
    - The classes generated MUST cover the # Input Data **AS MUCH AS POSSIBLE** and fall below the {{topic}} following **topic description**.
    {{max_class_number_requirement}}
    {{min_class_number_requirement}}
    - The response should be generated in json format following:
        {
            "classes": [
                {
                    "class_description" : <description_1>,
                    "class_name" : <title_1>
                },
                {
                    "class_description" : <description_2>,
                    "class_name" : <title_2>
                },
                <more classes...>
            ]
        }

Make sure output **pure json**

# Response
\end{minted}
\end{tcolorbox}
    \caption{Initial Taxonomy Generation Prompt For Subtopic}
    \label{fig:subtopic_initial}
\end{figure*}

\begin{figure*}[htbp]
    \centering
   \begin{tcolorbox}[colframe=borderblue, colback=bggray, left=1mm, right=1mm, top=0.5mm, bottom=0.5mm]
\begin{minted}{markdown}
# Context

You are a helpful assistant for clustering human-AI conversation within topic "{{topic}}". The following content in **Input Data** part are a batch of summarized human-AI conversation sampled. And a category table you generaeted based on the previous data in **# Category Table** part. You are going to update the table for downstream user interest discovery.

# Input Data

{{input_text}}

# Category Table

{{input_category_table}}

# Requirements

Your need to update the category table to make sure the table satisfy the following **requirements**:

    - The classes generated are the **TOPIC** of human-AI interaction MUST fall under the parent topic "{{topic}}". 
    - The parent topic description are as follows:
        {{topic_description}}
    - The class names and class descriptions generated can **ACCUREATELY** and **CONSISTENTLY** classify new data points into **1-3 class** with **NO AMBIGUITY**.



\end{minted}
\end{tcolorbox}
\end{figure*}

\begin{figure*}[htbp]
    \centering
   \begin{tcolorbox}[colframe=borderblue, colback=bggray, left=1mm, right=1mm, top=0.5mm, bottom=0.5mm]
\begin{minted}{markdown}
    - The class name should be a **CONCISE AND CLEAR** short sentence for the category.
    - The classes generated MUST be **MUTUAL EXCLUSIVE**.
    - The class description of each class should be generated within **200 WORDS** in English.
    - The class description MUST be generated based on data sample.
    - The class name must be consistent with its class description.
    - Output class must **fit the data as close as possible**, avoid adding unnecessary ones and missing necessary ones.    
    - Avoid general categories include any vague information such as "Other Topics", "Undefined", "Miscellaneous". 
    - You may ignore data points not related to {{topic}}.     
    - Keep each class **fine-grained**, AVOID include too many aspect in one class.
    - The classes generated MUST cover the # Input Data **AS MUCH AS POSSIBLE** and fall below the {{topic}} following **topic description**.
    {{max_class_number_requirement}}
    
# Instructions

You need to update using following steps: 

1. Review the given category table and the input data. Provide a rating score of current table. The rating score should between 0 to 100. The score should be given based instrinstic quality and extrinstic quality:

    - **Instrinstic quality**
        1) The categories meets the requirements given in ** # Requirements ** part, with clear and consistant category names and descriptions, and no overlap or contracdiction among the categories.
        2) The categories not include any vague information such as "Other Topics", "Undefined", "Miscellaneous". 
        3) Each category not contain too many aspects. 
        4) All categories are **MUTAL EXCLUSIVE**.
        5) The categories fall under the parent topic and adhere with topic description.

    - **Extrinstic quality**
        1) The data given can be classified into the 1-3 of given categories consistently without any ambiguity. 
        2) There is no missing category so that all new data can be classified properly.
        3) There is no unnecessary category that can be merged or removed.
        4) The categories are fine-grained and fit new data well.

2. Based on your score, decide if you need to update the categories, you can perform following operations:
    - Edit class name or class description of the categories.
    - Add new categories if there are missing categories.
    - Split one categories into multiple to become specific.
    - Merge multiple categories into one to become less amiguous. 
    - Remove unnecessary categories to reduce redundency.
    - No update if they are good enough.
    
    If you decide to update the categories, explain the update sugguestion in `suggesion` part. Otherwise just output `N/A` in suggestion part.

    Restate: The categories should be **CONCISE**, **CONSISTANT**, and **MUTAL EXCLUSIVE**. Make sure remember to update the dialogue count accordingly.

    Restate: Be **specific** about each category. **Do not include vague categories**

    You can ignore low quality or ambuiguous data points.


3. Output the report using json format as follows based on your decision and review result above, make sure categories satisfy the **requirements** given. 
    {
        "score": <table_score>,
        "suggestion: <suggestion>,
        "classes": [
            {
                "class_description" : <description_1>,
                "class_name" : <title_1>
            },
            {
                "class_description" : <description_2>,
                "class_name" : <title_2>
            },
             <more classes...>
        ]
    }

# Updated Category Table
\end{minted}
\end{tcolorbox}
    \caption{Taxonomy Update Prompt For Subtopic}
    \label{fig:subtopic_update}
\end{figure*}

\begin{figure*}[htbp]
    \centering
   \begin{tcolorbox}[colframe=borderblue, colback=bggray, left=1mm, right=1mm, top=0.5mm, bottom=0.5mm]
\begin{minted}{markdown}
# Context

You are a helpful assistant in analyzing user-AI interaction data. You are going to perform classification of user-AI interaction conversation based on a json version category table.

In the data given below, user requests starts with <User Request> and agent response starts wtih <Agent Response>. Utterance are separated by '----'. 

# Content

{{input_text}}

# Categories

{{input_categories}}


# Instruction 

You need to classify the given conversation and give confidence score of classification using the "conversation" field, "summary" field, # Categories table and given # Classification Examples with following requirements:

- You are classifying user-AI conversation under the topic of {{topic}}, the description of the the topic is:

    *topic description*

    {{topic_description}}

- Explain how you perform the classification in "explanation" part **WITHIN 300 WORDS**, cover both classification result and confidence score.

- All response should be in **ENGLISH**

- Classify based on the <User Request> in "conversation" , then refer to <Agent Response>, finally refer to "summary" if necessary.

- The classification MUST be done stick to "class_name" defined by "class_description".

- Perform classification ONLY FOCUS on the part related to {{topic}} and *topic description* of # Content. 

- You MUST classifiy the conversation into **AT MOST THREE** classes that are **HIGHLY RELEVANT**.

- The classification resulting label set MUST BE **AS SMALL AS POSSIBLE**, **HIGH PRECISION** and **COMPREHENSIVE**. 

- Respond the classes **ORDER BY RELEVANCE**, from most relevant to least relevant.

- "undefined" MUST not appear with other classes if there is any related turn or content.

- Give the relevance score correspond to each classification using an integer between 0-10. 

- Respond in **pure json** following with explanation and selected class **index** before the class name:
    {
        "explanation": <explanation>,
        "classes": [<class_index_1>, <class_index_2> ...],
        "relevance": [<relevance_1>, <relevance_2> ...]
    }

# Response
\end{minted}
\end{tcolorbox}
    \caption{Subtopic Assignment Prompt}
    \label{fig:subtopic_assignment}
\end{figure*}

\paragraph{Subtopic Label Assignment}
Finally, we label all data samples using the prompt illustrated in \Cref{fig:subtopic_assignment}, with the o3-mini model. For each topic, we select the best-performing taxonomy and use it to annotate all corresponding samples.
    
\subsection{Topic Label Quality Control}

After completing the labeling pipeline, we still observed some false positives upon manual inspection. To address this, we conducted an additional verification step—similar to the initial phase of the subtopic discovery pipeline—by reviewing each data sample alongside its raw conversation, assigned label, and label description, using the o3-mini model and the prompt shown in \Cref{fig:subtopic_verfication}. Following this verification, we removed all samples that lacked a valid label assignment or were assigned the Undefined label at either the topic or subtopic level. This filtering ensured that the final dataset aligned with the discovered taxonomy, ultimately reducing the dataset size to approximately 182k examples.

\begin{figure*}[htbp]
    \centering
   \begin{tcolorbox}[colframe=borderblue, colback=bggray, left=1mm, right=1mm, top=0.5mm, bottom=0.5mm]
\begin{minted}{markdown}
You are a careful classification data verifier, you are going to check multi-label classification of user-AI conversation result, you are going to check following conversation, the user request is start with <User Request>, and the AI response is start with <Agent Response>, the turns is separate by "----": 

# Conversation

{{input}}


# Classification Result

{{results}}

# Instruction

1. Carefully check if **each** classification result given in "class_description" under # Classification Result is highly relevant to the **major domain** of **any turn** of the conversation. 

2. Check class by class via verifying if any turn of conversation satisfy the "class_description", explain the result within 100 words after "explanation".

3. Respond json using following format, the "index" is the given index in # Classification Result and "check_result" is a string in "yes" or "no", choose yes if you are highly confident.

{
    "explanation": <explanation>,
    "results": [
        {
            "index": <label_index_as_int_1>,
            "check_result": <result_1>
        },
        {
            "index": <label_index_as_int_2>,
            "check_result": <result_2>
        },
        ...
    ]
}

# Response
\end{minted}
\end{tcolorbox}
    \caption{Subtopic Verification Prompt}
    \label{fig:subtopic_verfication}
\end{figure*}

\subsection{Keywords Categorization}

After the labeling process, we observed that certain topics—such as ``Fanfiction and Crossover'' and ``Programming'' contained a disproportionately large number of data samples. To enable more fine-grained question generation, we further categorized the extracted keywords into four semantic types: \textbf{programming language}, \textbf{creative artwork}, \textbf{public figure}, and \textbf{book}. Conversations that do not contain any keywords from these categories are classified as having no keywords.

\subsubsection{LLM Based Aggregation}

Assuming that the same word used by the same user conveys a consistent meaning, we first associate each user’s keyword with its corresponding description, extracted at the beginning of the process. We then employ o3-mini to cluster these raw keywords into semantically coherent groups, corresponding to categories including ``Programming Language'', ``Video Games'', ``Tabletop Games'', ``Manga/Anime'', ``Film'', ``TV Show'', ``Western Cartoon/Comic'', ``Book'', ``Musical'', and ``Public Figure'' , using the prompt illustrated in \Cref{fig:categorize_keywords}.

\begin{figure*}[htbp]
    \centering
   \begin{tcolorbox}[colframe=borderblue, colback=bggray, left=1mm, right=1mm, top=0.5mm, bottom=0.5mm]
\begin{minted}{markdown}
You are an expert in identifying the origin and clustering keywords with description, please complete following tasks

# Keywords

{{input}}

# Instruction

- You need to cluster **all keywords** and **keywords contained in description** given above via identifying all the **artwork, franchise, series, book, and public figures** it belong to like following results:

```json
   {
    "results": [
        {
            "name": "Doki Doki Literature Club!",
            "category": ["Video Games"],
            "keywords": ["Monika", "Natsuki", "Doki Doki Literature Club"]
        },
        {
            "name": "Game of Thrones",
            "category": ["TV Show"],
            "keywords": ["Daenerys Targaryen", "Arya Stark", "A Game of Thrones"]
        },
        {
            "name": "Dungeons & Dragons",
            "category": ["Tabletop Game"],
            "keywords": ["Dungeons and Dragons", "D&D", "DnD", "D&D 5e"]
        },
        <MORE EXAMPLES TRUNCATED TO SAVE SPACE ...>
        {
            "name": "Tom Holland",
            "category": ["Public Figure"],
            "keywords": ["Tom Holland", "tom holland"]
        },
        {
            "name": "Donald Trump",
            "category": ["Public Figure"],
            "keywords": ["Donald Trump", "Donald J. Trump"]
        }
    ]
}
```

- Descriptions of each keywords may lack information, you may need to **infer the underlaying artwork or franschise**.

- You need to copy the given keywords and keywords identified in "description" identically to "keywords" list in response.

- Respond empty list in "results" if there is no relatd artwork and media based on the category.

- You should ignore keywords that are not fall into any desired categories.

- You need to identify all artworks, series, franchise or book the given list of keywords belong to, use the **most well known and inclusive name**, and you respond without **detailed version or episode** using **English**

- **Avoid too general name**, such as DC Universe, Disney, Marvel Comics. **Focus on specific names**, such as Batmen, Spider-Man. '

- Public figure MUST be non-fictional people.

- Each unique public figure should have their own cluster with their most well-known name.

- You MUST focus on these categories only : "Video Games", "Tabletop Games", "Manga/Anime", "Film", "TV Show", "Western Cartoon/Comic", "Book", "Musical", and "Public Figure".

- You need to generate **no more than 80** results across all categories. Response most frequently referenced ones if more than 80.

- Respond **in pure json format** as the example above.

# Response

\end{minted}
\end{tcolorbox}
    \caption{Subtopic Verification Prompt}
    \label{fig:categorize_keywords}
\end{figure*}

\begin{figure*}[htbp]
    \centering
   \begin{tcolorbox}[colframe=borderblue, colback=bggray, left=1mm, right=1mm, top=0.5mm, bottom=0.5mm]
\begin{minted}{markdown}
You are a helpful assistant for translating structured data query over multi-lingual dataset into natural language for multiple choice question answering, the answer can have multiple correct options.

# Input

{{query}}

# Context

Explanation of condition fields:
    1. user_name: the unique user name of a user
    2. time_week: the start date of a week
    3. label_level_1: the topic or domain of a dialogue. 
    4. label_level_2: the subtopic or domain of a dialogue under a main topic in label_level_1.
    5. country: the country or region of the users' request come from. 
    6. language: the language the users are using. 
    7. keywords_aggregated: the keywords involved in the conversation, can be **one of** artworks/series/book/franchise, public figure and programming language.

# Examples

{{examples}}


# Instruction 
- The general idea of translation is to generate natural language question that **faithfully** describe the "condition" and ask about the "targat"
- You need to translate based on these condition explained in # Context. 
- The attribute used in question that describe keywords_aggregated options should be inferred from given target and options.
- You **MUST condense all description of topic or subtopic** in the generated question, using faithfully summarized version. 
- The question generated **MUST include all condition and target type** in **a natural and detailed way**.
- The question generated **MUST keep as much information as possible** from given topic description.
- Make sure the the generated question could be used as question of multiple choice question answering.
- Avoid leaking information and give hint in the question to the answer.
- Generate 2 possible questions with the same meaning but **diverse style**, **without target or candidate** in **English**, similar to proper # Examples.
- Respond in json format:
{
    "question_list": [<questions...>]
}

# Response
\end{minted}
\end{tcolorbox}
    \caption{Question Generation Prompt}
    \label{fig:generate_question}
\end{figure*}

\subsubsection{Rule-based LLM Result Aggregation}

Although o3-mini is prompted to generate the most well-known names for corresponding entities, the model occasionally produces inconsistent outputs, such as ``Pokémon'' vs. ``Pokemon''. These discrepancies are treated as distinct entries in downstream question generation. To address this, we define equivalence between a pair of large language model-generated terms or phrases $(w_a,w_b)$, where $\text{len}(w_a) <= \text{len}(w_b)$ -- based on a set of normalization criteria. Terms are considered equivalent across all keyword types except ``Public Figure'' if they satisfy any of the following conditions after applying string normalization: 

\begin{itemize}
    \item[1.] $w_a$ and $w_b$ are identical. 
    \item[2.] $w_a$ and $w_b$ are identical after removing all stopwords in NLTK English stopwords list. 
    \item[3.] $w_a$ is a prefix of $w_b$ and $w_a$ has more than 2 words. 
    \item[4.] $w_a$ is a suffix of $w_b$ and $w_a$ has more than 2 words.
    \item[5.] $w_a$ is an abbreviation of $w_b$ by concatenating all first letter of $w_b$.
\end{itemize}

For keywords of type ``Public Figure'' only Conditions 1 and 2 are applied due to the higher sensitivity of proper name matching. After normalization, we obtain a dataset with annotated two-level topic hierarchies and keywords spanning the following types: ``Programming Language'', ``Video Games'', ``Tabletop Games'', ``Manga/Anime'', ``Film'', ``TV Show'', ``Western Cartoon/Comic'', ``Book'', ``Musical'', and ``Public Figure''.

\subsection{Question Proposal} 

\paragraph{Attributes Combination} We generate questions through a brute-force search over various combinations and quantities of conditions. The full set of considered conditions is shown in \Cref{tab:att_intro}. Specifically, we enumerate all possible attribute combinations containing 0 to 3 conditions and manually select 73 meaningful combinations that can be naturally expressed in language. The selected combinations are listed in \Cref{tab:data_question_stat}.

\paragraph{Question Proposal Sampling} For each attribute condition and target type combination, we enumerate all possible condition value configurations using MongoDB. For each configuration, we first verify that the number of documents satisfying the condition is at least 50, unless the condition involves the username attribute, in which case the threshold is reduced to 10. This ensures that each generated question is supported by a sufficient number of documents.

Next, we query the database again to check whether the top 3 most frequent target attribute values collectively account for at least 15\% of all occurrences. This constraint prevents cases where the target distribution is overly uniform and lacks distinguishing signals.

All condition-target combinations that pass both checks are then stored in a map, where the key is the top-1 target value and the value is a list of corresponding condition-target combinations. Each list is sorted by the normalized entropy of the target distribution to prioritize more informative combinations.

Finally, we sample from this map in a round-robin manner, ensuring that each value is selected no more than twice. This strategy helps generate the most answerable questions while maintaining diversity across different top-1 target outcomes.

\subsection{Question Generation}
Given a set of condition types, corresponding values, and a target value, we prompt GPT-4.1 to generate natural language questions using the template shown in \Cref{fig:generate_question}.
\begin{figure}[htbp]
    \centering
   \begin{tcolorbox}[colframe=borderblue, colback=bggray, left=1mm, right=1mm, top=0.5mm, bottom=0.5mm]
\begin{minted}{markdown}
You are an helpful assistant in answering question about user-chatbot interaction in WildChat dataset.

# Conversations

{{conversations}}


# Question 

{{question}}

Base on the conversation given above, answer the given multiple choice question, **rank all options by relevance or correctness** based on the # Conversations. Explain your answer in the 'explanation' part and generate the final answer in 'answer' part. Respond using index of answer and using **pure json** format like:

{
    "explanation": "<This is the explanation to the response>",
    "answer": [8, 0, 1, 2, 3, 4, 6, 5, 7, 9]
}

# Answer
\end{minted}
\end{tcolorbox}
    \caption{Question Answering Prompt}
    \label{fig:question_answering}
\end{figure}

Following question generation, we retrieve the top 10 candidate answers for ranking by querying the database. In cases where fewer than 10 valid candidates are available, we supplement them by sampling from the global distribution of values that share the same target type.

Using this procedure, we generated a total of 6,177 questions.

\subsection{Question Quality Control}

We employ o4-mini for final quality control. Specifically, o4-mini is used to rank target candidates under two settings: (1) without any supporting context, and (2) with supporting context provided in the form of either summaries or raw conversations, using the prompting format shown in \Cref{fig:question_answering}.For each instance, we compute the instance-wise NDCG@10 score in the no-context setting, denoted as $s_{\text{no\_context}}$, and define the contextual score as $s_{\text{context}} = \max(s_{\text{raw\_context}}, s_{\text{summary\_context}})$, where $s_{\text{raw\_context}}$ and $s_{\text{summary\_context}}$are scores under raw and summarized contexts, respectively.

To assess statistical significance, we calculate a confidence-based threshold to determine whether a contextual improvement is meaningful over random performance. The threshold is defined as:
\begin{equation}
    s_{\text{threshold}} = \min(1.0, \max(0.0, s_{\text{random}} + z_{0.90} *  s_{\text{std}}))
\end{equation}

where $s_{\text{std}}$ is the standard deviation estimated via a Monte Carlo approach, and $z_{0.90}$ is the 90\%-confidence z-score.We remove any instance that satisfies both of the following conditions:
\begin{itemize}
\item $s_{\text{context}} - s_{\text{no\_context}} \le 0 $
\item $s_{\text{context}} < s_{\text{threshold}} $
\end{itemize}

After filtering, we retain a total of 6,027 valid data samples for downstream evaluation.

\end{document}